%% file: main.tex
\pdfoutput=1
\documentclass[11pt]{article}
\usepackage{amsmath, amsthm, amssymb, bbm, xparse, xspace}
\usepackage[margin=1in]{geometry}
\usepackage[round]{natbib}
\usepackage[dvipsnames,table]{xcolor}
\usepackage{hyperref}
\hypersetup{
    colorlinks=true,
    linkcolor=blue,
    citecolor=blue,
    urlcolor=blue,
    filecolor=blue
}
\usepackage[capitalise]{cleveref}
\crefname{subsection}{section}{sections}
\crefname{subsubsection}{section}{sections}
\crefname{assumption}{assumption}{assumptions}
\crefname{example}{example}{examples}
\usepackage{setspace}
\usepackage[shortlabels]{enumitem}
\usepackage{graphicx}
\usepackage{booktabs}
\usepackage{makecell}
\usepackage{listings}
\usepackage{float}
\usepackage{multirow}
\usepackage{colortbl}
\usepackage{etoolbox}
\usepackage[most]{tcolorbox}
\usepackage{algorithm}
\usepackage{algpseudocode}
\usepackage{tikz}
\usepackage{subcaption}
\usetikzlibrary{arrows.meta, positioning, shapes.geometric, fit, calc}

\usepackage[english]{babel}
\usepackage[autostyle, english = american]{csquotes}
\MakeOuterQuote{"}

\theoremstyle{plain}

\theoremstyle{definition}

\newtheorem{example}{Example}

\newcommand{\bE}{\mathbb{E}}

\newcommand{\bR}{\mathbb{R}}

\newcommand{\cS}{\mathcal{S}}
\newcommand{\cA}{\mathcal{A}}
\newcommand{\cX}{\mathcal{X}}

\newif\ifskeleton\skeletontrue
\ifskeleton
  \newcommand{\Note}[1]{\textcolor{red}{[\textbf{Note:} #1]}}
  \newcommand{\TODO}[1]{\textcolor{blue}{[\textbf{TODO:} #1]}}
  \newcommand{\jb}[1]{\textcolor{purple}{[\textbf{JB:} #1]}}
  
\else
  \newcommand{\Note}[1]{}\newcommand{\TODO}[1]{}\newcommand{\jb}[1]{}
  
\fi

\newcommand{\model}[1]{\texttt{#1}}          %
\newcommand{\sol}{\model{gpt-5.6-sol}\xspace}

\title{LLMs Can Design Near-Optimal OR Algorithms}

\author{Jackie Baek\thanks{Stern School of Business, New York University,
\href{mailto:baek@stern.nyu.edu}{baek@stern.nyu.edu}}}
\date{}

\makeatletter
\renewcommand{\paragraph}{\@startsection{paragraph}{4}{\z@}%
  {2.0ex \@plus .5ex \@minus .2ex}%
  {-1em}%
  {\normalfont\normalsize\bfseries}}
\makeatother

\begin{document}
\maketitle

\begin{abstract}
\input{sec_abstract}

\end{abstract}

\input{sec_introduction}

\input{sec_related_work}

\input{sec_setup}

\input{sec_inventory}

\input{sec_queueing}

\input{sec_assortment}

\input{sec_stability}

\input{sec_conclusion}

\section*{AI Assistance}
The author used Claude Code and Codex to help run the computational experiments
and help with writing the manuscript. The author takes full responsibility for
the content of the paper and any errors.

\bibliographystyle{abbrvnat}
\bibliography{bibliography}

\newpage

\appendix
\crefalias{section}{appendix}
\crefalias{subsection}{subappendix}
\crefalias{subsubsection}{subsubappendix}

\input{sec_appendix}

\end{document}

%% file: sec_abstract.tex
We ask whether large language models (LLMs) can design effective algorithms for well-specified operations research (OR) problems.
We study inventory control, queueing network control, and assortment optimization, using benchmark instances from recent papers in each area.
We evaluate two levels of LLM use: at level 1, the model receives one problem instance and returns a solution for that instance; at level 2, it receives only the problem class description and broad parameter ranges, and returns an algorithm that maps instance parameters to solutions.
Human input is minimal: we give one untuned prompt that
describes the problem, and the model has access to a Python sandbox tool with a fixed compute budget.

The strongest model we test, \model{gpt-5.6-sol}, matches or outperforms the best method reported in those papers on almost all evaluated instances.
This holds even at level 2, where the returned algorithm is fixed before seeing the evaluation instances.
Performance also improves sharply across models released less than eight months apart, suggesting
that this capability is moving quickly.
Thus, for the well-specified operations problems we study, a single untuned LLM query can already produce algorithms competitive with specialized methods.
These results suggest that frontier LLMs can be a serious empirical baseline for algorithm design in well-specified OR problems.

%% file: sec_introduction.tex
\section{Introduction}
\label{sec:intro}

Operations research (OR) has a long tradition of developing specialized algorithms for specific operational problems. Inventory control, queueing network control, assortment optimization, routing, and revenue management each have their own models, structural results, approximation methods, and computational heuristics. These algorithms are usually designed by researchers who use problem-specific knowledge to exploit the structure of a particular model or application domain.

Large language models (LLMs) are a new technology with broad capabilities. They can write code \citep{chen2021codex}, solve long-standing open math problems \citep{openai2026unitdistance}, and propose procedures across many domains \citep{romeraparedes2024funsearch}.
This raises a natural question:
\begin{center}
\emph{How good are LLMs at designing algorithms for operations problems?}
\end{center}

Answering this question can help clarify how LLMs may change the division of work in operations research. Solving a real operations problem involves many steps, including formulation, algorithm design, implementation, validation, and deployment. Each step requires different kinds of expertise and judgment \citep{lawless2025magicalbox}. We isolate the algorithm-design step and ask whether a general-purpose model can produce high-performing algorithms.

We study this question experimentally
on settings where the problem is mathematically well specified but exact optimization is difficult.
Given a formal description of the problem, we ask an LLM either to produce a solution for a particular instance or to design an algorithm that maps instance parameters to solutions.
We use the LLM in a simple one-query protocol: we provide one prompt that states the problem and the output format, without doing any prompt tuning or providing hints about the structure of a good solution.

We study three canonical OR domains with long literatures: inventory control, queueing network control, and assortment optimization. For inventory, we use the instances evaluated by \citet{gijsbrechts2022drl}: lost-sales systems with deterministic and stochastic lead times, dual sourcing, and multi-echelon distribution. For queueing, we use the multiclass queueing-network instances of \citet{dai2022queueing}: criss-cross networks, the N-model, and reentrant-line networks. For assortment, we use the hard benchmark instances of \citet{guo2026benchmark}: mixed-MNL, nested-logit, and constrained mixed-MNL choice models. Together, these benchmarks give 34 inventory instances, 13 queueing instances, and 3,393 assortment instances. A common feature of the source papers is that they use these instances to compare modern deep learning or reinforcement-learning approaches with classical methods.

We use an LLM in two ways, differing in where it enters the algorithm-design process:
\begin{itemize}
\item \textbf{Level 1:} the LLM is given a \textit{single instance} with its numeric parameters, and returns a
solution to that instance.
\item \textbf{Level 2:} the LLM is given a \textit{problem class} and broad parameter ranges. It
returns an algorithm that maps instance parameters to solutions, and that algorithm is then run on
every evaluation instance in the class. The returned algorithm is asked to produce each
instance-specific solution within 30 seconds.
\end{itemize}
For example, in assortment optimization, a level-1 query outputs one assortment for one specified choice model, whereas a level-2 query outputs an algorithm that maps choice-model parameters to an assortment.
Level 2 is closer to the standard notion of algorithm design; we also test level 1 because it is a strong and natural benchmark.
The methods developed in the papers whose instances we use are level-1 methods:
their policies are trained separately for each instance \citep{gijsbrechts2022drl,dai2022queueing,guo2026benchmark}.
For most evaluated instances, level 1 cannot rely on exhaustive enumeration due to the large solution space.

We evaluate four LLMs that span two providers and less than eight months of public release dates: \model{gpt-5.1}, \model{gpt-5.4}, and \model{gpt-5.6-sol} from OpenAI, and
\model{claude-fable-5} from Anthropic.
All code, prompts, and the complete run records (every LLM query with its reasoning summaries, executed code, outputs, and token usage) are publicly available.\footnote{\url{https://anonymous.4open.science/r/llm-or-algorithms-F9F2}.}

\subsection{Main findings}
\label{sec:intro_findings}

\begin{table}[!tp]
\centering
\input{tables/results_summary}
\caption{Summary of \model{gpt-5.6-sol}'s results against the best-performing method on that instance among those reported by the source paper.
Gain is the mean relative cost reduction (inventory and queueing) or revenue gain (assortment) over that method, in percent; positive means the LLM is better. No worse is the share of instances on which the LLM is within $0.1\%$ of or better than it.
The second column lists every method that is strictly the best on at least one instance of the class, and methods that tie the best on every instance.
Level 1 uses one query per instance; level 2 uses one algorithm per class, evaluated on every instance.}
\label{tab:summary}
\end{table}

\Cref{tab:summary} summarizes the results for the strongest model, \model{gpt-5.6-sol}, across all ten problem classes and both levels. Each entry compares the LLM with the best benchmark method \emph{for that instance}: in each domain, the best of the methods reported by the source paper on that instance, including the exact optimum where available.

\paragraph{\model{gpt-5.6-sol} matches or beats the best benchmark method on nearly every instance.}
Across the ten problem classes in \Cref{tab:summary}, \model{gpt-5.6-sol} has mean performance within $0.1\%$ of or better than the best benchmark method in every class at both levels.
It is no worse on \emph{every instance} in eight classes at level 1 and seven classes at level 2. This is a demanding comparison: the comparator is chosen instance by instance from the methods reported by the source paper, including exact dynamic programs or exact solvers where available.

\paragraph{Level 2 works well for structurally specified classes.}
Given a problem-class description and broad parameter ranges, a \emph{single}
level-2 query produces one algorithm for the entire class. The resulting
algorithms often perform nearly as well as level-1 solutions, which require
a separate query for each instance.
However, the level-2 approach can perform poorly when the problem class is too broad. When we used a single query for the full class of multiclass queueing networks, the resulting policy performed substantially worse than policies generated separately for each network type (\Cref{sec:q-results}).

\paragraph{The generated algorithms build on familiar OR ideas.}
The LLM outputs code which can be inspected and explained.
We find that the level-2 algorithms often follow familiar OR structures and use relatively simple rules.
For lost-sales inventory with deterministic lead times, the strongest algorithm extends an existing policy by adding a tuned standard-deviation term to the projected inventory level (\Cref{sec:inv-sol-l2}).
For reentrant queueing networks, the strongest algorithm is a variant of the weighted
maximum-pressure rule of \citet{dai2005pressure} (\Cref{sec:q-pressure-mechanism}). 
In assortment,
the strongest algorithms combine small exact routines, greedy and relaxation-based starts, and local improvement.

\begin{figure}[t]
    \centering
    \includegraphics[width=0.85\textwidth]{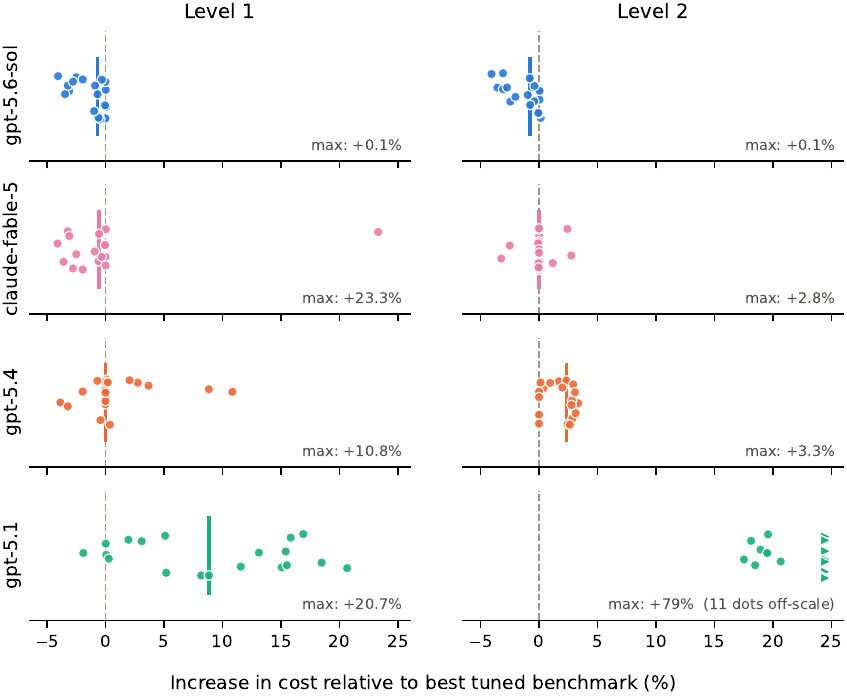}
    \caption{Preview of the inventory results on deterministic lost-sales instances (lower is
    better). Each dot is one instance, plotted as the percentage increase in cost relative to the
    best benchmark method on that instance (the best tuned benchmark policy, or the exact
    optimum on the six short-lead-time instances); negative values mean the LLM policy has lower
    cost. The vertical bar marks the median, and each panel reports its worst instance. Level 1
    uses one query per instance; level 2 uses one returned algorithm for the whole
    deterministic-lead-time family.
    Dots above $+25\%$ are clipped and shown at the right edge.
    }
    \label{fig:intro-lost-sales-det}
\end{figure}

    \paragraph{Performance improves with stronger models.}
The four models we evaluate were released within eight months of one another,\footnote{The public release dates are November 12, 2025
for \model{gpt-5.1}, March 5, 2026 for \model{gpt-5.4}, June 9, 2026 for
\model{claude-fable-5} (with access restored on July 1, 2026 after a temporary suspension), and
July 9, 2026 for \model{gpt-5.6-sol}, based on the providers' public release notes.
} yet their
performance differs substantially, especially at level 2. \Cref{fig:intro-lost-sales-det} gives a
preview of this pattern for deterministic lost-sales inventory: each dot is one instance, showing
the cost of the LLM policy relative to the best benchmark method on that instance, for all four
models at both levels.
This suggests that LLM
algorithm-design capability is moving quickly, and
that frontier LLMs are becoming a natural empirical baseline for hard OR problems.

\subsection{Implications and limitations}
\label{sec:intro_scope}

\paragraph{Implications for OR research.}
The strong LLM performance does not imply that LLMs replace algorithmic research. In the inventory and queueing case studies (\Cref{sec:inv-sol-l2,sec:q-pressure-mechanism}), the LLM adapts existing ideas, sometimes with a small performance gain. Neither case demonstrates a major new algorithmic principle. Therefore, developing new algorithmic structures and explaining why they work remain important goals for OR research. 

We cannot distinguish how much of the LLM performance depends on the existing literature, since we cannot test a model trained without it. In the cases we examine, however, the LLM adapts and tunes policy structures that already exist.  Whether LLMs can design equally strong algorithms for problem classes without an established algorithmic literature is an open question. 

Choosing a suitable policy structure and tuning it to the problem at hand are also valuable tasks in their own right. Our results suggest that LLMs can help with this, making established methods easier to apply. In settings where LLMs make algorithm design easier, researchers and practitioners can devote more attention to other stages of the OR workflow, including problem formulation, validation, and deployment.

\paragraph{Scope of the comparison.}
Our main comparisons use the methods that the source papers evaluated on these instances, and on
each instance we compare against the best of them. The source papers' authors chose these methods
to represent each setting, and they range from widely studied heuristics to exact solvers and deep
reinforcement learning, so we view them as a reasonable reference point. We do not claim that the
LLM algorithms beat \emph{every} method in the literature. When we looked for policies similar to the
LLM's algorithms in the case studies, we found some that the source papers did not evaluate, such
as projected inventory-level control for lost sales. Part of the LLM's advantage over the benchmark
methods may therefore come from using ideas that already exist in the literature but are less
widely used.

\paragraph{Empirical evidence and contamination.}
Our evidence is empirical and limited to three OR settings: inventory control, queueing network control, and assortment
optimization. A natural next step is to test the same framework on a wider range of operations
problems. We also cannot rule out that benchmark-specific information, such as instances, solutions,
or performance comparisons, appeared in model training. In
\Cref{sec:l2-holdout}, we partly address
this concern by evaluating the level-2 algorithms on new parameter values and generated instances
outside the main benchmarks.
Lastly, our results are empirical: we do not prove approximation
guarantees, and we do not know how robust the LLM-generated algorithms are outside the instance
families we test.
The queueing experiments also show that level-2 performance depends on how the problem class is defined: a class that is too broad can lead the model to return a generic policy that fails on some subclasses.

\paragraph{Roadmap.}
The next subsection reviews the related literature. \Cref{sec:setup} gives the shared experimental
setup and protocol used for all three domains.
\Cref{sec:inventory} presents the inventory experiments, \Cref{sec:queueing} presents the queueing
experiments, and \Cref{sec:assortment} presents the assortment experiments.
\Cref{sec:robustness} checks whether the main results survive removing sandbox compute,
extend to new instances, and are stable across repeated LLM queries. \Cref{sec:conclusion} concludes.

%% file: tables/results_summary.tex
\small\setlength{\tabcolsep}{2.5pt}
\begin{tabular}{llrrrrr}
\toprule[1.2pt]
 & & & \multicolumn{2}{c}{Level 1} & \multicolumn{2}{c}{Level 2} \\
\cmidrule(lr){4-5}\cmidrule(lr){6-7}
Problem class & Best benchmark methods & $n$ & Gain & No worse & Gain & No worse \\
& & & (\%) & (\%) & (\%) & (\%) \\
\midrule[1.2pt]
\multicolumn{7}{l}{\emph{Inventory (cost)}} \\
\quad \makecell[tl]{Lost sales,\\deterministic lead time} & \makecell[tl]{exact DP ($\ell \le 4$); tuned capped\\base stock; mixed strategy} & 19 & +1.32 & 100 & +1.27 & 94.7 \\[1pt]
\cmidrule(lr){1-7}
\quad \makecell[tl]{Lost sales,\\stochastic lead time} & tuned capped base stock & 7 & +1.55 & 85.7 & +0.60 & 100 \\[1pt]
\cmidrule(lr){1-7}
\quad Dual sourcing & exact DP & 6 & +0.00 & 100 & -0.03 & 83.3 \\[1pt]
\cmidrule(lr){1-7}
\quad \makecell[tl]{Multi-echelon\\distribution} & tuned constant order-up-to & 2 & +22.78 & 100 & +11.21 & 100 \\[1pt]
\midrule[1.2pt]
\multicolumn{7}{l}{\emph{Queueing (cost)}} \\
\quad Criss-cross & exact DP & 6 & +0.00 & 100 & +0.00 & 100 \\[1pt]
\cmidrule(lr){1-7}
\quad N-model & exact DP & 1 & +0.00 & 100 & +0.00 & 100 \\[1pt]
\cmidrule(lr){1-7}
\quad Reentrant line & \makecell[tl]{PPO trained per\\instance \citep{dai2022queueing}} & 6 & +5.50 & 66.7 & +5.00 & 83.3 \\[1pt]
\midrule[1.2pt]
\multicolumn{7}{l}{\emph{Assortment (revenue)}} \\
\quad MMNL & NN + local search; ADXOpt & 628$^{*}$ & +0.00 & 100 & +0.00 & 100 \\[1pt]
\cmidrule(lr){1-7}
\quad Nested logit & \makecell[tl]{LP policy of \citet{kunnumkal2023};\\ADXOpt; NN + local search} & 971$^{*}$ & +0.00 & 100 & +0.00 & 100 \\[1pt]
\cmidrule(lr){1-7}
\quad Constrained MMNL & \makecell[tl]{conic MIP;\\NN + local search} & 1794$^{*}$ & +0.00 & 100 & +0.00 & 100 \\[1pt]
\bottomrule[1.2pt]
\end{tabular}
\vspace{2pt}\par\footnotesize $^{*}$Level 2 is evaluated on every instance of the class; level 1 needs one query per instance and is evaluated on a stratified subset of 72 instances for MMNL, 48 instances for Nested logit, 36 instances for Constrained MMNL.

%% file: sec_related_work.tex
\subsection{Related Work}
\label{sec:related}

\paragraph{LLMs for operations research and operational decision-making.}

A growing literature studies LLMs in operational decision-making.
We organize these works by where the LLM enters the decision pipeline: formulation, supplying model inputs, designing the solution, and making the decision.
Surveys of the area span these stages and emphasize reliability,
modeling errors, and tool use \citep{orsurvey2025, simchilevi2026chapter}.
\citet{simchilevi2026democratizing} describe how generative AI helps practitioners use existing
optimization models.

\paragraph{1. Formulating the model.} One stream treats formulation as a translation problem: given
natural-language problem statements, LLMs produce optimization models, solver code, or
model files \citep{ramamonjison2023nl4opt, ahmaditeshnizi2024optimus, orlm2024, zhou2025dpbench}.
Other work builds benchmarks, training data, and search procedures for the same formulation task
\citep{astorga2025autoformulation, huang2024mamo, yang2025optibench, jiang2024llmopt,
zhang2024optllm}.
Another line of work studies messier, more realistic settings in which the model is built, repaired, or specified interactively rather than translated from a clean problem statement
\citep{xiao2024coe, li2023optiguide, ao2026optirepair, drossman2026interactive, lawless2024iwant,
liang2026largescale}.

\paragraph{2. Constructing model inputs.} A second stream uses LLMs to construct or elicit the primitives
an optimization model needs after the model class has been chosen. \citet{baek2026llmsaa} use
LLM-generated personas to build distributions for sample-average approximation,
\citet{huangwang2026virtual} use LLM-powered virtual populations for demand simulation and pricing,
and \citet{duan2025askclarify} study human--LLM clarification of inventory-control inputs.

Our experiment deliberately removes these first two stages. The optimization problem is already
specified in precise mathematical terms, so we do not test formulation. Every parameter is known
exactly and stated in the prompt, so we do not test data generation.

\paragraph{3. Designing the solution method.} 
A third stream uses LLMs to design solution methods, where the problem formulation is given as input.
Our paper belongs in this stream.
A line of work embeds the LLM in an iterative algorithm-discovery loop. In inventory, \citet{huang2026invevolve} use evolutionary search for white-box policies, and related systems repeatedly generate, evaluate, and revise code or heuristic ideas \citep{romeraparedes2024funsearch, liu2024eoh, ye2024reevo, vanstein2024llamea, zheng2025mctsahd, alphaevolve2025, yang2024opro}; see \citet{liu2026llm4ad} for a survey. 
\citet{kim2026autoopt} use a domain-specific pipeline to design and verify first-order optimization methods.
Our goal is different from these works; we do not try to optimize the scaffold around the LLM. Instead, we ask how well a frontier LLM performs under a simple single-query protocol: the same prompt structure, sandbox, and compute budget are used across problem classes.

A related line uses LLMs to design components of a prespecified solution framework.
In stochastic optimization, \citet{zhang2026genso} use an LLM to generate basis functions
for recourse decision rules. In mixed-integer optimization, \citet{lawless2025coldstart}
configure cutting-plane separators from a problem description and solver documentation, while other
work uses LLM-guided search to generate cuts, large-neighborhood-search rules, or branching
policies \citep{yazdani2025evocut, ye2025llmlns, hou2026llm4branch}. In our level-2 experiments,
the LLM chooses the overall structure of the solution method.

\paragraph{4. Direct decision-making.}
A fourth stream places the LLM directly in the decision seat and measures its actions in inventory, retail, economic, pricing, and supply-chain environments \citep{baek2026agents, llmnewsvendor2025, aimbench2025, tanlamai2026housing, cohenhage2026confirmation, ahmed2026pricing, long2025beergame}. Related work studies LLM exploration in bandit and economic environments whose specification the model must learn through interaction \citep{krishnamurthy2024explore, fish2025econevals}. Our level-1 experiments are related, especially in assortment, where the model returns a decision rather than a reusable procedure. 
The key difference is that, in much of the existing work, the LLM still has room for subjective judgment about formulation: what objective to pursue and which constraints matter.

\paragraph{General methods for OR algorithm design.}
Before LLMs, the main general-purpose alternative to problem-specific algorithm design was machine learning and reinforcement learning. This literature has produced learned policies and value-function approximations for inventory control \citep{vanroy1997,gijsbrechts2022drl,temizoz2025dcl,xie2026deepstock,alvo2023hdpo,harsha2021parl}, queueing control \citep{moallemi2008approximate,shah2020stable,qu2020scalable,liu2022rl,wei2024sample,dai2022queueing,chen2024qgym}, and assortment optimization and routing \citep{wang2023neuralchoice,aouad2022deepchoice,li2026assortmentdrl,bello2017neural,kool2019attention}.
Recent work uses a transformer architecture and reinforcement learning to learn policies for joint replenishment \citep{liu2026ortransformer}.
We ask a similar question with LLMs in place of learned policies, and unlike these black-box methods, the LLM returns an algorithm written as readable code that can be inspected and analyzed (see case studies in \Cref{sec:inv-sol-l2,sec:q-pressure-mechanism}).

\paragraph{Benchmarks and baselines.}

Our experiments use published benchmark families in inventory control, queueing-network
scheduling, and assortment optimization.
This follows recent calls for more systematic empirical benchmarking in stochastic OR:
\citet{dong2026empirical} argue for evaluating algorithms on structured instance families, using
reusable simulators, shared baselines, and explicit protocols rather than isolated numerical
examples. 
Recent benchmarks for LLM-driven algorithm design, such as CO-Bench \citep{sun2026cobench} and HeuriGym \citep{chen2026heurigym}, also ask LLM agents to write reusable algorithms. 
FrontierOR \citep{kong2026frontieror} asks LLMs to formulate large-scale optimization problems from prose descriptions and design algorithms to solve them. In comparison, we isolate algorithm design by providing the full mathematical specification.

Our inventory instances are based on \citet{gijsbrechts2022drl} and cover lost sales with
deterministic and stochastic lead times, dual sourcing, and multi-echelon distribution.
For lost-sales inventory, our benchmarks include myopic, base-stock, constant-order, and capped
base-stock policies \citep{huh2009lostsales, goldberg2016constant, johansen2008restricted, xin2021, xin2026capped}.
For dual sourcing, we use dual-index \citep{veeraraghavan2008}, tailored base-surge
\citep{allon2010,xin2018tbs}, and capped dual-index \citep{sun2019cdi} policies. 
The case study in \Cref{sec:inv-sol-l2} compares the LLM's rule with projected inventory-level
(PIL) control \citep{vanjaarsveld2024pil}, which is outside the main benchmark set.

The queueing benchmark of \citet{dai2022queueing} covers multiclass networks, where the controller decides which queue each server works on. Exact dynamic programming is feasible for the small criss-cross and N-model instances but not for the larger reentrant lines, where \citet{dai2022queueing} compare per-instance reinforcement learning with structured scheduling rules. The reentrant-line instances also appear in QGym \citep{chen2024qgym}, and our case study in \Cref{sec:q-pressure-mechanism} relates the LLM's rule to the maximum-pressure policies of \citet{dai2005pressure}.

The assortment benchmark of \citet{guo2026benchmark} covers three settings: mixed multinomial
logit (MMNL) with a cardinality constraint \citep{bront2009, rusmevichientong2014random};
nested logit with or without per-nest cardinality constraints
\citep{davis2014nested, gallego2014constrained, kunnumkal2023}; and MMNL with general linear
constraints \citep{sen2018conic}.

%% file: sec_setup.tex
\section{Setup}
\label{sec:setup}

We begin by setting up the framework used throughout the experiments: a problem class, its
instances, and the feasible solutions for each instance (\Cref{sec:primitives}). We then formalize
the different granularities at which an LLM can be queried to solve such problems
(\Cref{sec:levels}).

\subsection{Problem Classes and Instances}
\label{sec:primitives}

A \emph{problem class} is a tuple $C = (\Theta, \cX, R)$. Here $\Theta$ is a parameter
space, and a \emph{problem instance} is a specific parameter $\theta \in \Theta$. For each instance,
$\cX(\theta)$ is the set of feasible solutions, and $R(\theta, x)$ is a deterministic reward function for solution $x \in \cX(\theta)$ on instance $\theta$. An instance may contain randomness,
such as demand realizations or customer choices; $R$ folds this randomness into a single
deterministic quantity, for example through an expectation or a long-run average.

The elements of $\cX(\theta)$ may be simple or structured objects, and this is the only
distinction we need between single-shot and sequential problems. In a \emph{single-shot} problem, the
solution is a single choice, such as an assortment. In a \emph{sequential} problem, the
decision-maker observes a state and acts repeatedly, so a solution is a policy that selects an
action at each state,
$\cX(\theta) \;=\; \{\, \pi : \cS \to \cA \,\}$,
where $\cS$ and $\cA$ are the state and action spaces.

We illustrate the primitives with two examples.

\begin{example}[Inventory control with lost sales] \label{ex:inv_lostsales}
The problem class $C$ is single-product lost-sales inventory control. An instance is
\[
\theta \;=\; (F, \ell, h, p, c, \bar q),
\]
where $F$ is the demand distribution, $\ell$ is the deterministic lead time,
$h$ is the holding cost, $p$
is the lost-sales penalty, $c$ is the unit ordering cost, and $\bar q$ is the maximum order quantity.
The problem is sequential.
The state is on-hand inventory together
with the pipeline of outstanding orders, $s = (I, q_1, \ldots, q_{\ell-1}) \in \cS$, and an
action is an order quantity $a \in \cA = \{0, 1, \ldots, \bar q\}$. Thus the feasible-solution
set $\cX(\theta)$ is the policy space $\{\pi : \cS \to \cA\}$. Each period, the order placed
$\ell$ periods earlier arrives, a new order is placed, demand $d_t \sim F$ is realized, and unmet
demand is lost. Writing $I_t$ for on-hand inventory after arrivals, the
reward of a policy is the negative long-run average cost,
\[
R(\theta, \pi) \;=\; -\lim_{T \to \infty} \frac{1}{T}\,
\bE\!\left[\, \sum_{t=1}^{T} h\,(I_t - d_t)^{+} + p\,(d_t - I_t)^{+} + c\,a_t \right],
\]
where $a_t$ is the order placed in period $t$.
\end{example}

\begin{example}[Assortment optimization under MMNL] \label{ex:assortment_mmnl}
The problem class $C$ is assortment optimization under the mixed multinomial logit (MMNL)
choice model with a cardinality cap. An instance is
\[
\theta \;=\; (m, n, \omega, u, v_0, r, k),
\]
where $m$ is the number of customer segments with weights $\omega \in \Delta_m$, $n$ is the
number of products, $u \in \bR_{\ge 0}^{m \times n}$ are utilities, $v_0 \in \bR_{> 0}^{m}$ are
outside-option weights, $r \in \bR_{\ge 0}^{n}$ are prices, and $k$ is a cardinality cap. The
problem is single-shot, and the solution space is
$\cX(\theta) = \{ S \subseteq [n] : |S| \le k \}$. The reward is expected revenue: a customer in
segment $j$ offered assortment $S$ buys product $i \in S$ with probability
$u_{ji} / (v_{0j} + \sum_{i' \in S} u_{ji'})$, so
\[
R(\theta, S) \;=\; \sum_{j=1}^{m} \omega_j \,
\frac{\sum_{i \in S} r_i\, u_{ji}}{v_{0j} + \sum_{i \in S} u_{ji}}.
\]
\end{example}

How broad to make a problem class is itself a modeling choice. In Example~\ref{ex:assortment_mmnl} we defined the class as assortment optimization under MMNL, so an instance with
a different choice model belongs to a different class. One could instead define the broader class of assortment optimization under an
arbitrary choice model; that class contains the one in Example~\ref{ex:assortment_mmnl}.
We use this flexibility in the queueing experiments by trying two definitions of level 2: a primary definition with fixed queueing-network structures, and a broader problem class definition that allows arbitrary multiclass networks up to the same size range.

\subsection{Levels of LLM invocation}
\label{sec:levels}

\begin{figure}[H]
\centering
\begin{tikzpicture}[
  font=\small,
  llm/.style  = {draw, rounded corners=2pt, fill=blue!14, minimum height=6.5mm,
                 minimum width=16mm, align=center, inner sep=2pt},
  arr/.style  = {-{Latex[length=1.8mm]}},
  hd/.style   = {font=\small},
  chain/.style= {draw=black!65, rounded corners=2pt, fill=black!8, font=\small\bfseries,
                 minimum height=6.5mm, inner sep=4pt},
  chainarr/.style = {-{Latex[length=2.4mm,width=2mm]}, line width=0.9pt, black!65},
  guide/.style= {black!12, dotted, line width=0.6pt},
  tail/.style = {draw=black!35, rounded corners=3pt, fill=black!3, inner sep=6pt,
                 align=left, font=\footnotesize, text width=88mm}
]
\def\xR{1.75}
\def\xC{5.0}  \def\xT{7.7}  \def\xX{10.4}
\def\yA{0}    \def\yB{-1.3}
\foreach \x in {\xC,\xT,\xX} \draw[guide] (\x,1.05) -- (\x,-1.75);
\draw[black!25] (\xR+1.95,1.15) -- (\xR+1.95,-1.75);
\node[chain] (hC) at (\xC,1.45) {class $C$};
\node[chain] (hT) at (\xT,1.45) {instance $\theta$};
\node[chain] (hX) at (\xX,1.45) {solution $x$};
\draw[chainarr] (hC) -- (hT);  \draw[chainarr] (hT) -- (hX);
\node at (\xR,\yA) {returns algorithm $\sigma$};
\node[llm] (a2) at (\xC,\yA) {LLM$(C)$};
\node      (s2) at (\xT,\yA) {$\sigma(\theta)$};
\node      (x2) at (\xX,\yA) {$x$};
\draw[arr] (a2) -- (s2);  \draw[arr] (s2) -- (x2);
\node at (\xR,\yB) {returns solution $x$};
\node[llm] (a1) at (\xT,\yB) {LLM$(\theta)$};
\node      (x1) at (\xX,\yB) {$x$};
\draw[arr] (a1) -- (x1);
\node[anchor=east, align=right] at (\xR-1.9,\yA) {\textbf{Level 2}};
\node[anchor=east, align=right] at (\xR-1.9,\yB) {\textbf{Level 1}};
\draw[arr, black!55] (\xX,\yB-0.35) -- (\xX,-2.2);
\node[tail, anchor=north] at (\xT,-2.2)
  {\textbf{single-shot} (assortment): $x$ is the offered set of products.\\[2pt]
\textbf{sequential} (inventory, queueing): $x$ is a policy, a mapping from the state
   to an action.};
\end{tikzpicture}
\caption{ A level is the point in the chain
at which the LLM is queried. Level 2 is
queried once per class and returns an algorithm that serves every instance; level 1 is queried once
per instance and returns a solution for that instance alone.
}
\label{fig:levels}
\end{figure}
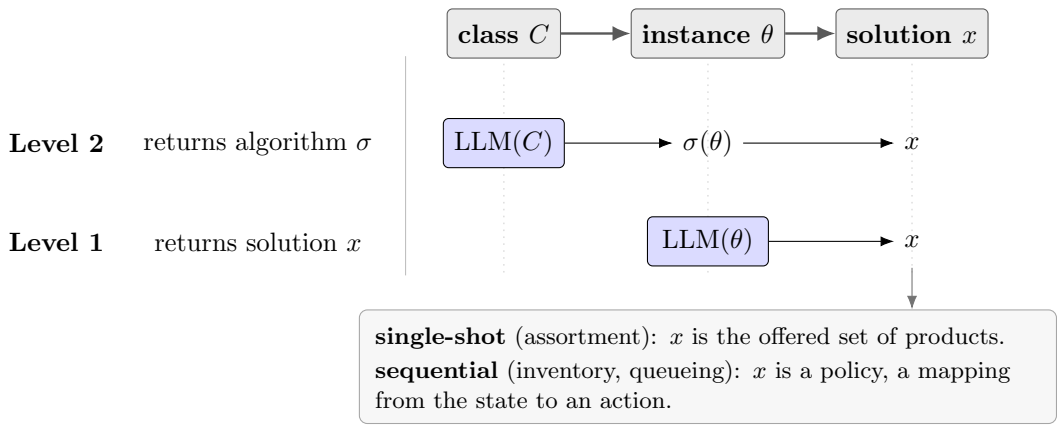

A problem class induces a chain of objects, shown in \Cref{fig:levels}, and an LLM can be inserted at different points in this chain.
We call the choice of insertion point the \emph{level of invocation}:

\begin{itemize}
\item \textbf{Level 1.} The LLM is queried once per instance: it receives
$\theta$ and returns a \emph{solution} $x \in \cX(\theta)$.
\item \textbf{Level 2.} The LLM is queried once per class: it receives a
description of $C$ and returns an \emph{algorithm} $\sigma$, executable code that maps any
instance $\theta$ of the class to a solution $\sigma(\theta) \in \cX(\theta)$.
\end{itemize}

We use \emph{artifact} for the concrete code returned by one LLM query. This applies at both
levels: a level-1 artifact implements one instance-specific solution, while a level-2 artifact
implements a reusable algorithm.

These are the two levels our experiments use. For sequential problems the chain runs one step
further, since a solution is a policy and the decision-maker acts at every state, and a third
level becomes available:

\begin{itemize}
\item \textbf{Level 0.} The LLM is queried at every state: it receives
$(\theta, s_t)$ and returns an \emph{action} $a_t$.
\end{itemize}

We do not test level 0 in this paper; however, other papers in the literature operate at this
level (see \Cref{sec:related}).

\paragraph{Resource constraints.}
Without further restrictions, the levels can collapse into one another. A level-2 algorithm
could simply be $\sigma(\theta) = \mathrm{LLM}(\theta)$.
We therefore put resource constraints on both the query and the returned executable object, and we
disallow LLM calls from returned code entirely.
One could define budgets that permit LLM access, or a combination of LLM access and code execution,
but we do not study those variants in this paper.

Note that a lower level has strictly more information: a model queried at level 1 sees the instance itself, while a level-2 algorithm can only execute its written code on it.
However, this informational advantage of level 1 does not necessarily translate into better empirical performance;
our experiments compare performance across the levels.

\subsection{Experimental protocol}
\label{sec:protocol}

Each query is a single tool-using session:
the model receives a mathematical problem description, may use a Python sandbox under a
prompt-stated compute budget, and returns the required deliverable.
The sandbox provides the Python standard library, \texttt{numpy}, and \texttt{scipy}, but no external optimization solvers such as Gurobi.

\begin{figure}[H]
\centering
\begin{tikzpicture}[
  font=\small,
  >=Stealth,
  shared/.style={
    draw=black!45,
    fill=black!4,
    rounded corners=2pt,
    align=left,
    inner sep=6pt,
    text width=0.52\textwidth
  },
  branch/.style={
    draw=black!50,
    fill=white,
    rounded corners=2pt,
    align=left,
    inner sep=6pt,
    text width=0.38\textwidth
  },
	  tag/.style={font=\bfseries, align=center}
	]
\node[font=\bfseries\large, text=black!80] at (0,1.15) {Prompt schema example};

\node[shared] (problem) at (0,0) {
\textbf{Describe the lost-sales inventory problem.}

Level 1 gives numeric parameters; level 2 gives parameter names and broad ranges of the parameters.
};

\node[shared, below=0.45cm of problem] (sandbox) {
\textbf{Describe Python tool and computation budget.}

The model has 3600 seconds of total Python compute, which it can divide across up to 50 calls.
};

\node[branch, below left=1.1cm and 0.45cm of sandbox.south] (lone) {
\textbf{Describe output.}

Python code defining the function
\[
\texttt{def order(on\_hand, pipeline):}
\]
which returns an integer order quantity in \texttt{[0, max\_order]}.
};

\node[branch, text width=0.39\textwidth, below right=1.1cm and 0.45cm of sandbox.south] (ltwo) {
\textbf{Describe output.}

Python code defining the function
\[
\texttt{def design(params):}
\]
where \texttt{design} receives the instance parameters and returns an
\texttt{order(on\_hand, pipeline)} function.
};

\draw[->, thick, black!60] (problem) -- (sandbox);
\draw[->, thick, black!60] (sandbox.south) -- ++(0,-0.35) -| (lone.north);
\draw[->, thick, black!60] (sandbox.south) -- ++(0,-0.35) -| (ltwo.north);
\node[font=\bfseries, anchor=east, text=black!65] at ($(lone.north)+(0,0.36)$) {Level 1};
\node[font=\bfseries, anchor=west, text=black!65] at ($(ltwo.north)+(0,0.36)$) {Level 2};
\end{tikzpicture}
\caption{Example schematic of the lost-sales inventory prompt.
}
\label{fig:inv-prompt-schematic}
\end{figure}
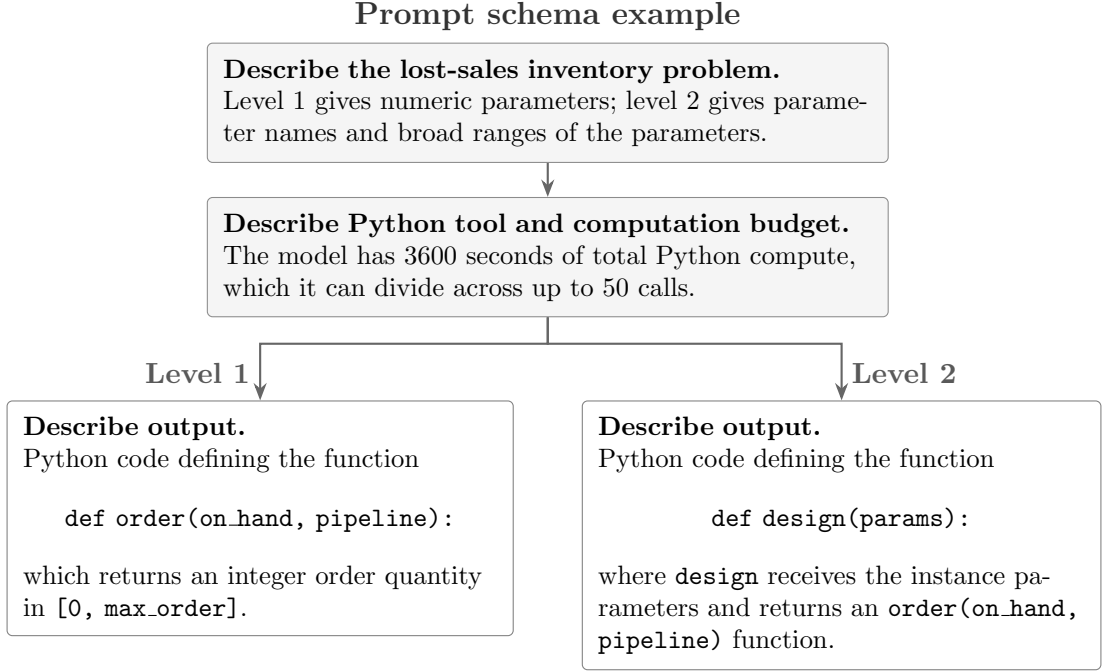

The prompts are deliberately untuned. They state the problem, the available sandbox, the compute budget, and the required output format, but do not name benchmark policies, suggest solution methods, or describe the shape of a good answer. Returned code is evaluated outside the LLM session and may not call an LLM. \Cref{fig:inv-prompt-schematic} gives an abbreviated schematic for the lost-sales inventory prompts. Representative prompt examples are reported in \Cref{app:prompts}; the full prompt texts are in the public repository.

In the main results, each LLM query is run once: once for each model-instance pair at level 1
and once for each model-class pair at level 2. We do not otherwise retry or select over repeated
samples, except when an artifact is defective; the rerun rule is described in \Cref{app:protocol}.
\Cref{sec:l2-stability} checks how much the results vary across repeated runs of the same query.
We compare every LLM result with the \emph{best benchmark method on the same
instance}: the best reported or implemented method from the source paper, including exact solvers where available.

%% file: sec_inventory.tex
\section{Inventory Experiments}
\label{sec:inventory}

Our inventory experiments follow the experiments of \citet{gijsbrechts2022drl}, who asked
whether a single deep reinforcement learning method can match specialist inventory heuristics.
We keep their instances and benchmarks where possible, changing only the cases that would otherwise
be exactly solvable.

\subsection{Instances}
\label{sec:inv-instances}

We use the three inventory settings in \citet{gijsbrechts2022drl}: lost sales, dual sourcing, and
multi-echelon distribution. \Cref{tab:inv-instances} lists the instance groups.
The main text focuses on the 26 lost-sales instances, and we defer the dual-sourcing and multi-echelon descriptions and results to \Cref{app:additional-inventory}.

\begin{table}[H]
\centering
\small
\resizebox{\textwidth}{!}{%
\begin{tabular}{llll}
\toprule
\begin{tabular}[l]{@{}l@{}}\textbf{Problem class}\\\textbf{(\# instances)}\end{tabular} & \textbf{Group} & \textbf{Specification} & \begin{tabular}[l]{@{}l@{}}\textbf{Optimum}\\\textbf{reachable?}\end{tabular} \\
\midrule
\multirow{4}{*}{\begin{tabular}[l]{@{}l@{}}Lost sales,\\deterministic\\(19)\end{tabular}}
& Short lead times & $\ell \in \{2,3,4\} \times p \in \{4,9\}$, $\lambda = 5$ & Yes \\
& Lead-time sweep & $\ell = 5, \ldots, 10$, $\lambda = 5$, $p = 4$ & No \\
& Demand sweep & $\lambda \in \{10, 15, 20, 25\}$, $p = 4$, at $\ell = 8$ & No \\
& Penalty sweep & $p \in \{9, 19, 49\}$, $\lambda = 5$, at $\ell = 8$ & No \\
\midrule
\begin{tabular}[l]{@{}l@{}}Lost sales,\\stochastic\\(7)\end{tabular}
& &
\begin{tabular}[l]{@{}l@{}}
$\ell \sim U\{2, \ldots, \bar\ell\}$, $\bar\ell = 5, \ldots, 11$ \\
$\lambda = 5$, $p = 4$
\end{tabular} & No \\
\midrule
\begin{tabular}[l]{@{}l@{}}Dual sourcing\\(6)\end{tabular}
& &
\begin{tabular}[l]{@{}l@{}}
$\ell_r \in \{2,3,4\}$, $c_e \in \{105,110\}$ \\
demand $U\{0,\ldots,4\}$
\end{tabular} & Yes \\
\midrule
\begin{tabular}[l]{@{}l@{}}Multi-echelon\\(2)\end{tabular}
& &
\begin{tabular}[l]{@{}l@{}}
$(\ell_w,\ell_r,\mu,\sigma) \in \{(2,2,5,14),(5,3,0,20)\}$ \\
10 stores
\end{tabular} & No \\
\bottomrule
\end{tabular}}
\caption{Inventory problem classes and instance groups from \citet{gijsbrechts2022drl}. The main text focuses on the
lost-sales groups; details for dual sourcing and multi-echelon distribution are in
\Cref{app:additional-inventory}.}
\label{tab:inv-instances}
\end{table}

The first four lost-sales groups use the deterministic-lead-time class in \Cref{ex:inv_lostsales}: a single location orders from one supplier, demand is Poisson, unmet demand is lost at penalty $p$, and orders are constrained by a per-period maximum order quantity. We also include a separate stochastic-lead-time lost-sales class, where each order's lead time is random. All lost-sales instances share holding cost $h=1$, ordering cost zero, and maximum order quantity $\bar q=\lceil 2.5\lambda\rceil+2$. At short deterministic lead times, the exact optimum is reachable within a query's own compute budget: value iteration solves the largest of those six instances in about ten seconds. The remaining deterministic instances test whether the model constructs a good heuristic.

\paragraph{Deviations from the source paper.}
For lost sales, we move the demand and penalty sweep groups to $\ell = 8$, rather than $\ell = 4$ that was used in \citet{gijsbrechts2022drl}.
At $\ell = 4$, every instance in those two sweeps is still exactly solvable, so they would mostly repeat the short-lead-time test.

\subsection{LLM experiments and benchmark policies}
\label{sec:inv-llm}
\label{sec:inv-benchmarks}

We query LLM models at both level 1 and level 2. 
Level 1 uses one query per instance; level 2
uses one query per problem class, with deterministic and stochastic lead times treated as separate
classes. 
Each query has a sandbox budget of 3600 seconds of Python compute across at most 50
calls. The returned code is evaluated outside the LLM session: level-1 code may do one-time setup
at module import, and level-2 code defines a \texttt{design} routine that maps instance parameters to
an ordering policy. In both cases, the prompt asks this setup or design step to finish within 30
seconds per instance, and asks the per-period \texttt{order} function to be lightweight. Additional details are reported in \Cref{app:protocol}; prompt examples are reported in \Cref{app:prompts}.

\paragraph{Benchmark policies.}
No simple policy is optimal in any of these settings, so the literature has produced structured
 heuristics, several of them carrying performance guarantees.
We write $I$ for on-hand inventory at the moment of ordering and $Q$ for the total quantity on
order but not yet arrived, so the inventory position is $I + Q$, and $(z)^+ = \max\{z,0\}$.
The quantities shown in parentheses are the policy's free parameters; for each benchmark policy, the free parameters are tuned separately for each instance by grid search.
\begin{itemize}[leftmargin=*, itemsep=2pt, topsep=3pt]
\item \emph{Base-stock} $(S)$: order $(S - I - Q)^+$. Optimal when unmet demand is
backordered, but suboptimal under lost sales.
\item \emph{Constant-order} $(r)$: order $r$ every period, whatever the state. Introduced by
\citet{reiman2004} and asymptotically optimal as the lead time grows
\citep{goldberg2016constant}, since a long pipeline leaves the current state progressively less
informative about what will be needed when an order finally arrives.
\item \emph{Capped base-stock} $(S, b)$: order $\min\{\,b,\ (S - I - Q)^+\}$, first introduced by \citet{johansen2008restricted} as a restricted base-stock policy; this policy class has a $2.33$-approximation guarantee \citep{xin2021,xin2026capped}.
\item \emph{Mixed strategy} $(r, P)$: order $r$ with probability $P$ and $r + 1$ otherwise.
Introduced by \citet{gijsbrechts2022drl}, after observing their trained A3C policy alternate
between two order sizes, because order quantities are integers while the best constant rate
generally is not.
\item \emph{Myopic} (no free parameters): order to minimize expected cost over the next one or two
periods.
\end{itemize}

Simulator validation and scoring details are in \Cref{app:protocol,app:validation}.

\subsection{Results}
\label{sec:inv-results}

We summarize the main lost-sales results here;
the full results are reported in \Cref{app:full}.

\begin{figure}[t]
    \centering
    \includegraphics[width=0.85\textwidth]{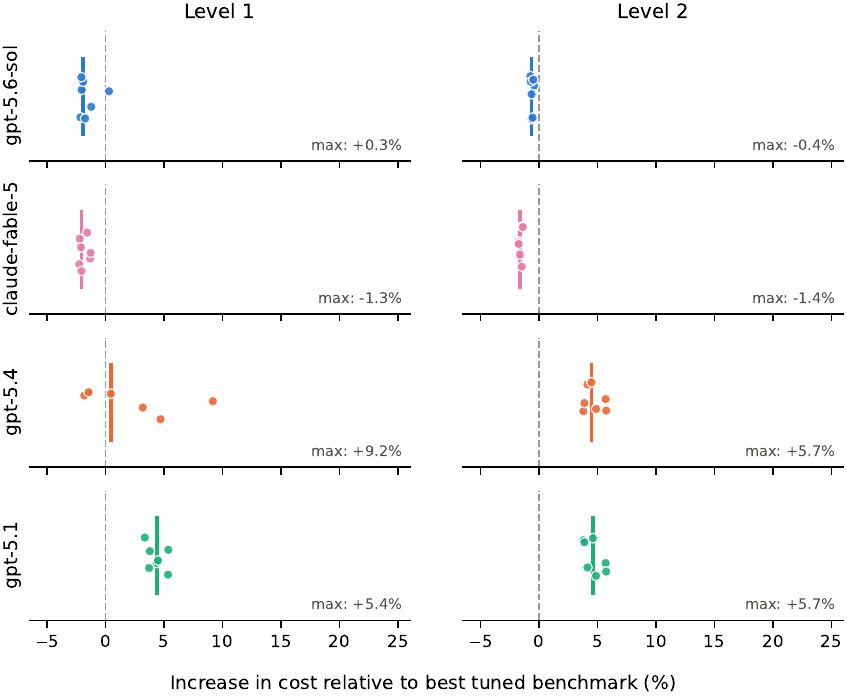}
    \caption{Analog of \Cref{fig:intro-lost-sales-det} for stochastic-lead-time instances.
    Lost-sales performance on stochastic-lead-time instances (lower is better). Each dot is one instance, plotted as
    the percentage increase in cost relative to the best tuned benchmark policy on that instance;
    negative values mean the LLM policy has lower cost. The vertical bar marks the median, and each
    panel reports its worst instance.
    Level 1 uses one query per instance; level 2 uses one returned algorithm for the
    stochastic-lead-time family.
    Dots above $+25\%$ are clipped and shown at the right edge.}
    \label{fig:lost-sales-stoch}
\end{figure}

\paragraph{Overall performance.}
\Cref{fig:intro-lost-sales-det,fig:lost-sales-stoch} give the main lost-sales result, split by
deterministic and stochastic lead times. Each dot is measured against the best benchmark method on that instance. 
Across all 26 instances, \model{gpt-5.6-sol} is within $0.5\%$ of the best benchmark method at both levels. 
On the 20 lost-sales instances beyond the exactly solvable grid, \model{gpt-5.6-sol} has lower simulated cost than the best tuned benchmark on 19 instances at level 1 and all 20 at level 2. 
At level 2, the model writes
only two algorithms, one for deterministic lead times and one for stochastic lead times.

\model{claude-fable-5} shows that the strong performance is not unique to one provider, but it is less reliable than \model{gpt-5.6-sol}. Across the 26 lost-sales instances, it is within $0.5\%$ of the best tuned benchmark on 25 instances at level 1 and 23 at level 2.
The weaker OpenAI models show a clear capability gradient. \model{gpt-5.4} is competitive at level 1 but never beats the tuned benchmark at level 2, while \model{gpt-5.1} is worse at both levels. Since the four models were released within less than eight months, these gaps suggest rapid progress. 

For deterministic lead times, the level-2 algorithm comes from one class-level query:
598 seconds of wall time, including 54.9 seconds of sandbox compute. After the query, the returned
algorithm's design time is 3.0 seconds per instance on average, with a maximum of
4.8 seconds. By comparison, level 1 uses one query per deterministic instance; across the same
instances, the mean query wall time is 2{,}248 seconds, including 1{,}959 seconds of sandbox
compute, and the mean module-load time is 1.8 seconds. Full inventory timing results are in
\Cref{tab:inventory-timing-l2,tab:inventory-timing-l1}.

We now examine selected instance groups more closely to understand where these aggregate results
come from.

\paragraph{Exactly solvable short-lead-time instances.}
\begin{table}[t!]
\centering
\input{tables/results_tier_a_llm}
\caption{The short lead-time grid (exact optimum computed): LLM policies, level~1 (L1) and level~2 (L2) per model. On these instances the best benchmark method is the exact dynamic program, so entries are the simulated cost percentage gap above the exact optimum (first column); \textbf{bold} marks a simulated cost gap within $0.1\%$ of the optimum. All 95\% CI half-widths are below 0.05.}
\label{tab:full-tier-a}
\end{table}

Table~\ref{tab:full-tier-a} focuses on the six short-lead-time instances.
They are small enough that dynamic programming can compute the exact optimum, and the table reports gaps above it.
A bolded number denotes a simulated cost gap within $0.1\%$ of the optimum.
This table is a diagnostic for whether the model recognizes when exact optimization is feasible.
\model{gpt-5.6-sol} and \model{claude-fable-5} recover the optimum at level 1 and remain essentially optimal at level 2. The weaker models are less consistent, especially at level 2. Thus, when exact optimization is within reach, the strongest models are able to compute it.

\begin{figure}[t!]
\centering
\includegraphics[width=0.8\textwidth]{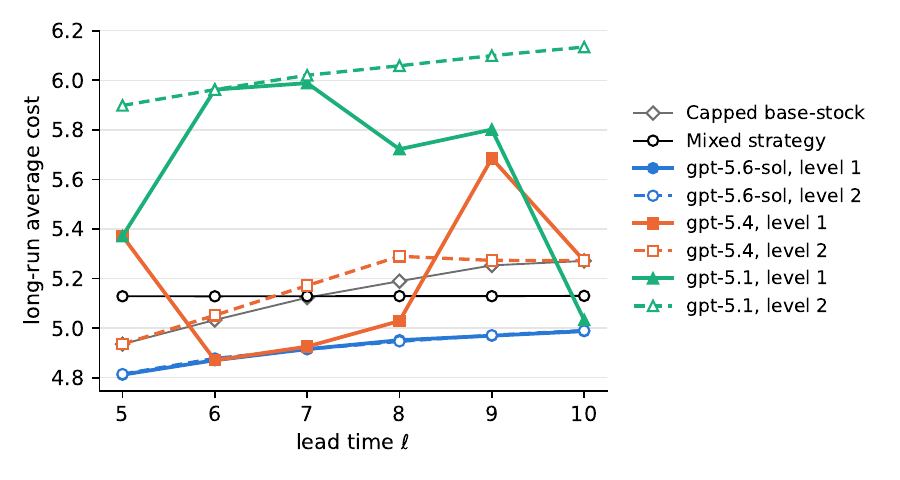}
\caption{Analogue of Figure~5 in \citet{gijsbrechts2022drl}: deterministic lead-time sweep with
Poisson demand mean $\lambda=5$ and penalty $p=4$.
The thin solid lines are the two strongest tuned benchmarks: capped base-stock (grey diamonds)
and the mixed strategy (black circles); the mixed strategy becomes the better of the two at long
lead times. Solid lines with filled markers
are level-1 policies, one query per instance; dashed lines with open markers are each model's
single deterministic level-2 algorithm evaluated on the same instances. We show the three OpenAI
models here to keep the figure readable; \model{claude-fable-5} appears in
\Cref{fig:intro-lost-sales-det} and the appendix tables.}
\label{fig:model-comparison}
\end{figure}

\paragraph{Lead-time sweep.}
\Cref{fig:model-comparison} zooms in on the deterministic lead-time sweep, where exact dynamic
programming is no longer practical.
We plot the performance of \emph{Capped base stock} and \emph{Mixed strategy}, the benchmark policies that performed the best on this family.
\model{gpt-5.6-sol} improves on
the best tuned benchmark at every lead time by $2.5\%$ to $4.1\%$, and the two curves nearly overlap.
This figure shows the model-capability gradient in a more concrete way. \model{gpt-5.4} is competitive at level 1 on several lead times, but its level-2 algorithm stays close to the capped base-stock benchmark rather than improving on it.

\paragraph{Comparison with A3C.}
In the lost-sales experiments of \citet{gijsbrechts2022drl}, A3C mostly matches strong heuristics rather than clearly improving
on them. 
On the short-lead-time grid, A3C does not find the optimum and is outperformed by capped base-stock.
On the deterministic lead-time sweep, A3C initially outperforms the standard benchmarks, but this observation motivates
the mixed strategy, which then matches the A3C performance. 
In contrast, \model{gpt-5.6-sol}
improves on the best tuned benchmark in \Cref{fig:model-comparison}, including the mixed strategy,
at every lead time.

In our taxonomy, A3C is a level-1 approach:
the training and hyperparameter tuning are done separately for each instance, producing a policy for
that instance rather than an algorithm that transfers across instances. This pipeline is expensive:
the paper reports that one hyperparameter setting takes about 24 CPU-hours to evaluate, and that
the automatic tuning procedure evaluates approximately 250 hyperparameter settings. Our level-2 algorithms come
from one query with at most one hour of sandbox compute, plus the light setup computation reported
in \Cref{sec:query-time-compute}. The returned object is also different. A3C produces a black-box
policy, whereas the level-2 LLM algorithm is inspectable and interpretable code.

\input{sec_inventory_mechanism}

%% file: tables/results_tier_a_llm.tex
\small\setlength{\tabcolsep}{4.5pt}
\begin{tabular}{lrrrrrrrrr}
\toprule
 & Optimum & \multicolumn{2}{c}{\model{gpt-5.6-sol}} & \multicolumn{2}{c}{\model{claude-fable-5}} & \multicolumn{2}{c}{\model{gpt-5.4}} & \multicolumn{2}{c}{\model{gpt-5.1}} \\
Instance & & L1 & L2 & L1 & L2 & L1 & L2 & L1 & L2 \\
\midrule
$\ell=2,\,p=4$ & 4.395 & \textbf{+0.02} & \textbf{+0.04} & \textbf{+0.02} & \textbf{+0.02} & \textbf{+0.02} & +1.70 & \textbf{+0.02} & +40.04 \\
$\ell=2,\,p=9$ & 6.094 & \textbf{+0.07} & +0.14 & \textbf{+0.07} & \textbf{+0.07} & \textbf{+0.07} & +2.85 & \textbf{+0.07} & +79.01 \\
$\ell=3,\,p=4$ & 4.599 & \textbf{-0.00} & \textbf{+0.01} & \textbf{-0.00} & \textbf{-0.00} & \textbf{-0.00} & +2.34 & +8.18 & +33.91 \\
$\ell=3,\,p=9$ & 6.531 & \textbf{-0.01} & \textbf{+0.03} & \textbf{-0.01} & \textbf{-0.01} & \textbf{-0.01} & +2.93 & +5.10 & +36.50 \\
$\ell=4,\,p=4$ & 4.729 & \textbf{-0.05} & \textbf{-0.03} & \textbf{-0.05} & \textbf{-0.05} & \textbf{-0.05} & +2.00 & +0.28 & +30.40 \\
$\ell=4,\,p=9$ & 6.836 & \textbf{-0.01} & \textbf{+0.05} & \textbf{-0.01} & \textbf{-0.01} & \textbf{-0.01} & +3.33 & +1.95 & +37.60 \\
\bottomrule\end{tabular}

%% file: sec_inventory_mechanism.tex
\subsection{A mechanism case study: the LLM inventory policy}
\label{sec:inv-sol-l2}

We examine the structure of the main level-2 algorithm from \model{gpt-5.6-sol} for deterministic lost sales and relate it to existing inventory policies. We then assess how much its new algorithmic component contributes to performance.

\paragraph{The LLM algorithm.}
For deterministic lead times, the main level-2 algorithm returned by
\model{gpt-5.6-sol} projects usable stock through the lost-sales dynamics until
just before the new order arrives. Let $\hat m_t$ and $\hat v_t$ denote its
approximations to the mean and variance of that remaining stock. These moments
are computed recursively using a normal approximation with truncation at zero
(details in \Cref{app:inventory-mechanism}). The algorithm orders
\begin{equation}
q_t^{\mathrm{LLM}}=\min\{\bar q,\,(T-\hat m_t-\gamma\sqrt{\hat v_t})^+\},
\label{eq:inventory-uncertainty}
\end{equation}
rounded to an integer. Here $\bar q$ is the physical order limit, $T$ is a
stock target, and $\gamma$ controls the uncertainty adjustment. The
\texttt{design} routine tunes $T$ and $\gamma$ by simulation, first over a
coarse grid and then around promising candidates. A negative $\gamma$ adds a
buffer that grows with projected standard deviation.

\paragraph{Relationship to existing policies.}
Many inventory policies take the form
\begin{equation}
q_t=\min\{c,\,(T-P_t)^+\},
\label{eq:inventory-template}
\end{equation}
where $c$ is an order cap, $T$ is a stock target, and $P_t$ is an effective
inventory position. The LLM algorithm takes this form with $c=\bar q$ and
$P_t=\hat m_t+\gamma\sqrt{\hat v_t}$. Existing policies use different choices
of $c$ and $P_t$. Including the physical order limit $\bar q$ in the cap,
and leaving integer rounding implicit:
\begin{itemize}
\item \textbf{Base stock:} $c=\bar q$ and $P_t=I_t+Q_t$, where $I_t$ is
on-hand stock and $Q_t$ is the total outstanding quantity. There is no fitted
cap beyond the physical limit.
\item \textbf{Capped base stock:} $c=\min\{b,\bar q\}$ and
$P_t=I_t+Q_t$, where $b$ is a fitted order cap.
\item \textbf{Projected inventory-level (PIL) control:} $c=\bar q$ and
$P_t=m_t=\bE[J_t]$, where $J_t$ is usable stock just before the new order
arrives, conditional on the current state. This projection accounts for
scheduled arrivals and intervening lost demand.
\end{itemize}
Projected-mean targeting appears in \citet{sun2016}, PIL is analyzed by
\citet{vanjaarsveld2024pil}, and an extension is used in the pharmaceutical
inventory study of \citet{stranieri2025pil}.
PIL was not among the lost-sales comparators evaluated by
\citet{gijsbrechts2022drl}, whose benchmark set we retain for the main results.

Using the same approximation to $m_t$, setting $\gamma=0$ in the LLM rule
recovers PIL. Thus, the extension relative to PIL is the 
\emph{separately tuned standard-deviation term}. When $\gamma<0$, the LLM rule uses
$P_t=\hat m_t-|\gamma|\sqrt{\hat v_t}$: at the same projected mean, it orders
more when future stock is more uncertain.

\paragraph{Performance impact of the uncertainty term.}
To measure this extension's contribution, we compare separately fitted PIL and
two-parameter policies on all $19$ deterministic settings. We refit PIL's
$T$ with $\gamma=0$ and the extended rule's $(T,\gamma)$ under the same
projection, simulation-search framework, rounding, and physical cap, then
evaluate on fresh common demand paths. These subsequent fits study the policy
family; they do not replace the original LLM scores. Calibration, projection,
and inference details are in \Cref{app:inventory-mechanism}.

\begin{figure}[t]
\centering
\includegraphics[width=0.90\textwidth]{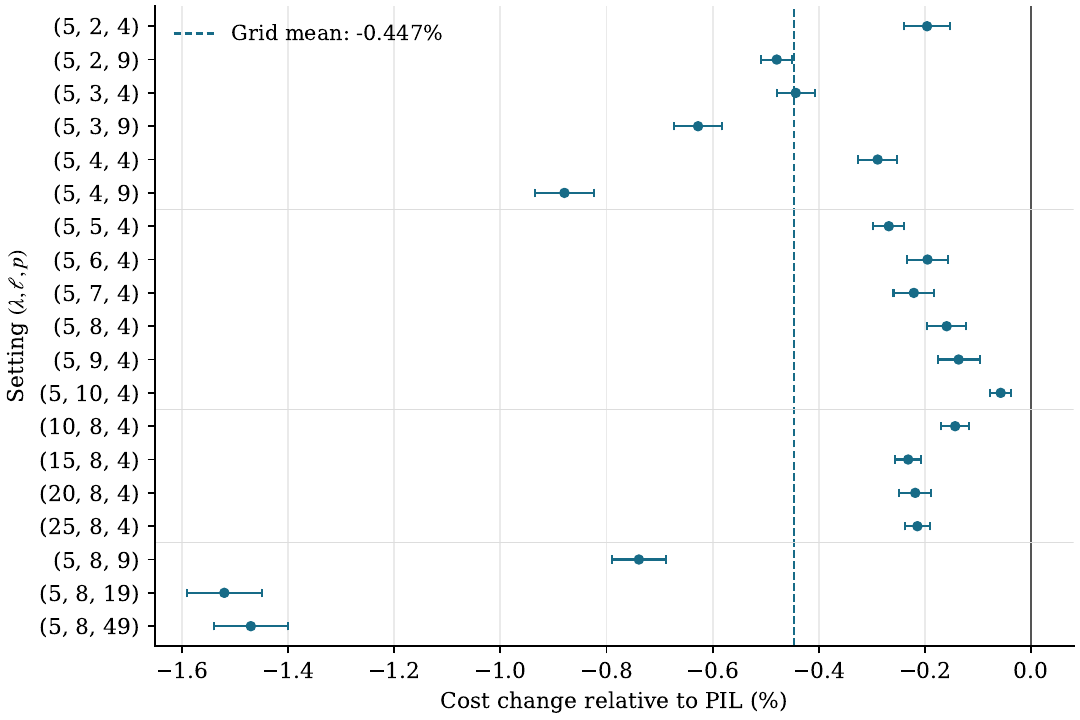}
\caption{Cost change from adding the standard-deviation coefficient to PIL,
using the original moment approximation and separately fitted parameters.
Points show $100(C_{\mathrm{two\text{-}parameter}}/C_{\mathrm{PIL}}-1)$ with
paired pointwise 95\% confidence intervals; negative values favor the
extended rule.}
\label{fig:inventory-pil-mechanism}
\end{figure}

\Cref{fig:inventory-pil-mechanism} shows the result. Adding the standard-deviation
term lowers cost relative to PIL on all $19$ settings, by $0.447\%$ on average
(95\% confidence interval $[0.438\%,0.456\%]$). The fitted $\gamma$ is negative
in every setting, so the extended rule orders more when projected stock is more
uncertain, as a safety buffer would. 
Therefore, the new standard-deviation term gives a small but meaningful improvement over PIL on these instances.

%% file: sec_queueing.tex
\section{Queueing Experiments}
\label{sec:queueing}

Our queueing experiments use the multiclass queueing-network instances of
\citet{dai2022queueing}, who study whether deep reinforcement learning can produce strong
scheduling policies for queueing networks.

\subsection{Instances}
\label{sec:q-instances}

In a multiclass queueing network, jobs move through a sequence of queues, and a few shared servers
must decide which queue to work on at each moment.
Formally, there are $Q$ queues, one per job class, and $S$ servers. Jobs arrive from outside to
queue $q$ as a Poisson process with rate $\lambda_q$, which may be zero. Server $s$ can work on
queue $q$ only if $A_{sq}=1$, and then completes jobs at exponential rate $\mu_{sq}$. When a job
finishes service at queue $q$, it either leaves the network or joins a fixed downstream
queue as a job of that class. A single job can therefore pass through several classes, possibly
returning to a server it has already visited.
Whenever a job arrives or finishes service, the controller sees the vector of queue lengths and
assigns each server to a compatible nonempty queue or leaves it idle. Several servers can work on
the same queue at once, each on a different job, and a server can be pulled off a job in progress,
which later resumes where it left off.

Each job waiting in queue $q$ costs $h_q$ per unit time. Let $X_q(t)$ be the number of jobs in
queue $q$ at time $t$. The goal is to minimize the long-run average holding cost
\[
\limsup_{T\to\infty} \frac{1}{T}
\mathbb{E}\left[\int_0^T \sum_{q=1}^Q h_q X_q(t)\,dt\right].
\]

\Cref{tab:q-instances} summarizes the 13 instances. The criss-cross and N-model instances are low-dimensional enough for dynamic programming. 
The reentrant-line networks, which \citet{dai2022queueing} call ``extended six-class networks'', have $3L$ queues for $L=2,\ldots,7$ stations and are not tractable
by dynamic programming at these sizes; for these instances, the strongest existing comparator is the PPO policy
reported by \citet{dai2022queueing}, which is trained separately for each instance.

\begin{table}[H]
\centering
\begin{tabular}{llll}
\toprule
\textbf{Group} & \textbf{Instances} & \textbf{Size} & \textbf{Best benchmark method} \\
\midrule
Criss-cross & 6 &
2 servers, 3 queues
& DP optimum \\
N-model & 1 & 2 servers, 2 queues & DP optimum \\
Reentrant line & 6 & 2--7 servers, 6--21 queues & per-instance PPO \\
\bottomrule
\end{tabular}
\caption{Queueing instance groups from \citet{dai2022queueing}; detailed instance parameters are in \Cref{app:q-instances}. The reentrant-line networks are also used in QGym \citep{chen2024qgym}.}
\label{tab:q-instances}
\end{table}

\subsection{LLM experiments and benchmark policies}
\label{sec:q-llm}
\label{sec:q-benchmarks}

\paragraph{Two level-2 class definitions.}
We report two level-2 settings. The main L2 setting uses one query for each structural class: criss-cross networks, the N-model, and reentrant-line networks.  We also run one Broad L2 prompt for the full multiclass-network class.

Level 1 uses one query per instance; level 2 uses one query per class. Each query has a sandbox budget of 3600 seconds of Python compute across at most 50 calls. The returned level-2 \texttt{design} routine is asked to finish within 30 seconds per instance, and the returned event-level scheduling policy is asked to be lightweight.

\paragraph{Benchmark policies.}
For the seven DP-solvable instances, the benchmark is the DP-computed optimum computed by us. The computed
optima reproduce the published values of \citet{dai2022queueing} to three or four digits on the
criss-cross instances, with the largest heavy-traffic instance sensitive to truncation. For the
reentrant-line networks, the primary benchmark is the PPO policy of \citet{dai2022queueing}.
We also re-simulate classical static policies such as LBFS and $c\mu$ under our steady-state
protocol as secondary checks. Simulator validation and comparator details are reported in
\Cref{app:queueing}.

On the seven instances where dynamic programming is feasible, we compute each returned policy's long-run average cost and compare it with the DP-computed optimum. 
For the reentrant-line instances, we estimate long-run average costs by simulation.
See \Cref{app:q-validation} for details.

\subsection{Results}
\label{sec:q-results}

\Cref{tab:q-main-results} reports the queueing results for \model{gpt-5.6-sol}. We report cost
gaps relative to the best benchmark method on the same instance, so negative values mean the LLM
policy has lower cost.

\begin{table}[H]
\centering
\small
\input{tables/results_queueing_main}
\caption{Queueing performance of
\model{gpt-5.6-sol}. Entries are percentage cost gaps relative to the best benchmark method; negative
is better. The L2 column uses one query per structural subclass, while Broad L2 uses one query for the full
queueing class.
Bold entries beat the comparator or are within $0.1\%$ of the optimum.
}
\label{tab:q-main-results}
\end{table}

\paragraph{Level 1.}
At level 1, the model is essentially optimal on all seven DP-solvable instances.
Thus, when exact dynamic programming is feasible, the model usually
discovers and uses it.
On the larger reentrant-line networks, the comparator is PPO trained separately for each network. The
LLM beats PPO on four of six instances, is essentially tied on one, and is worse on the largest
network by $2.2\%$, within PPO's reported uncertainty.

\paragraph{Level 2.}
The main L2 setting preserves most of the L1 performance. For criss-cross and the N-model, the
returned algorithms solve truncated dynamic programs and are essentially optimal. For the
reentrant-line class, one class-level query returns a simulation-tuned structural policy that beats PPO on five of six instances.

\paragraph{Broad L2.}
We also ran a Broad L2 query that described the whole multiclass queueing-network model up to
7 servers and 21 queues. This is much broader than the structural classes used for the main L2 queueing results.
The Broad L2 query returns a generic index-plus-pressure rule. It is competitive on the reentrant-line networks, but performs poorly on the criss-cross and N-model instances, where good
policies require topology-specific threshold behavior.
These results show that the definition of the problem
class matters: a structurally specified class lets the model choose an appropriate strategy, while a
too-broad class can elicit a reasonable-looking generic rule with classical failure modes.

\paragraph{Results across models.}
\Cref{tab:queueing_models} summarizes the queueing results for all four models as the
average percentage cost gap to the best benchmark method over the instances of each class.
At level 1,
all four models are within $1\%$ of optimal on the small classes (except \model{gpt-5.1} on the
N-model) and beat PPO on average on the reentrant lines. At level 2 the models separate:
\model{gpt-5.6-sol} and \model{claude-fable-5} stay essentially optimal on the criss-cross class, while only \model{gpt-5.6-sol} improves on the existing methods for the reentrant lines.
\model{gpt-5.1}'s reentrant-line algorithm
reduces to a static priority rule, and its N-model algorithm returns policies that fail to
stabilize the system.

\begin{table}[!tp]
\centering
\small
\caption{Average percentage cost gap to the best benchmark method on each queueing
class (negative means LLM is better).
Level 2 uses the structurally specified classes. Entries that match or beat the best benchmark method (within $0.1\%$) are in
bold.}
\label{tab:queueing_models}
\input{tables/results_queueing_models}

\smallskip
    {\footnotesize $^{\dagger}$\model{gpt-5.1}'s N-model algorithm
returns policies that do not stabilize the system.\par}
\end{table}

\input{sec_queueing_mechanism}

%% file: tables/results_queueing_main.tex
\begin{tabular}{llrrr}
\toprule
\textbf{Class} & \textbf{Instance} & \textbf{L1} & \textbf{L2} & \textbf{Broad L2} \\
\midrule
\multirow{6}{*}{\makecell[l]{Criss-cross\\ \emph{vs.\ exact optimum}}}
 & imbalanced, light & $\mathbf{0.0\%}$ & $\mathbf{0.0\%}$ & $+2.8\%$ \\
 & balanced, light & $\mathbf{0.0\%}$ & $\mathbf{0.0\%}$ & $+3.9\%$ \\
 & imbalanced, medium & $\mathbf{0.0\%}$ & $\mathbf{0.0\%}$ & $+13.7\%$ \\
 & balanced, medium & $\mathbf{0.0\%}$ & $\mathbf{0.0\%}$ & $+22.2\%$ \\
 & imbalanced, heavy & $\mathbf{+0.001\%}$ & $\mathbf{+0.006\%}$ & $+22.9\%$ \\
 & balanced, heavy & $\mathbf{+0.001\%}$ & $\mathbf{+0.002\%}$ & $+82.7\%$ \\
\midrule
\makecell[l]{N-model\\ \emph{vs.\ exact optimum}}
 & $\rho=0.95$, $h=(3,1)$ & $\mathbf{0.0\%}$ & $\mathbf{+0.003\%}$ & $+98.2\%$ \\
\midrule
\multirow{6}{*}{\makecell[l]{Reentrant line\\ \emph{vs.\ per-instance PPO}}}
 & $L=2$ stations, $6$ queues & $\mathbf{-1.9\%}$ & $+1.1\%$ & $+9.5\%$ \\
 & $L=3$ stations, $9$ queues & $\mathbf{-8.6\%}$ & $\mathbf{-6.5\%}$ & $+6.7\%$ \\
 & $L=4$ stations, $12$ queues & $\mathbf{-8.7\%}$ & $\mathbf{-8.5\%}$ & $+4.4\%$ \\
 & $L=5$ stations, $15$ queues & $+0.5\%$ & $\mathbf{-6.3\%}$ & $+2.0\%$ \\
 & $L=6$ stations, $18$ queues & $\mathbf{-16.6\%}$ & $\mathbf{-9.8\%}$ & $\mathbf{-2.7\%}$ \\
 & $L=7$ stations, $21$ queues & $+2.2\%$ & $\mathbf{-0.1\%}$ & $\mathbf{+0.1\%}$ \\
\bottomrule
\end{tabular}

%% file: tables/results_queueing_models.tex
\begin{tabular}{llcccc}
\toprule
Level & Class & \model{gpt-5.6-sol} & \model{claude-fable-5} & \model{gpt-5.4} & \model{gpt-5.1} \\
\midrule
\multirow{3}{*}{1} & Criss-cross ($n=6$) & {\boldmath$+0.0\%$} & {\boldmath$+0.0\%$} & {\boldmath$+0.1\%$} & $+0.9\%$ \\
 & Reentrant line ($n=6$) & {\boldmath$-5.5\%$} & {\boldmath$-9.0\%$} & {\boldmath$-4.3\%$} & {\boldmath$-1.4\%$} \\
 & N-model ($n=1$) & {\boldmath$+0.0\%$} & {\boldmath$+0.0\%$} & {\boldmath$+0.0\%$} & $+4.1\%$ \\
\midrule
\multirow{3}{*}{2} & Criss-cross ($n=6$) & {\boldmath$+0.0\%$} & {\boldmath$+0.0\%$} & $+3.8\%$ & $+8.4\%$ \\
 & Reentrant line ($n=6$) & {\boldmath$-5.0\%$} & $+0.2\%$ & $+2.6\%$ & $+5.8\%$ \\
 & N-model ($n=1$) & {\boldmath$+0.0\%$} & $+1.4\%$ & $+0.3\%$ & unstable$^{\dagger}$ \\
\bottomrule
\end{tabular}

%% file: sec_queueing_mechanism.tex
\subsection{A mechanism case study: the LLM reentrant-line policy}
\label{sec:q-pressure-mechanism}

We take a closer look at \model{gpt-5.6-sol}'s level-2 reentrant-line algorithm to understand how it relates to existing policies.

\paragraph{The LLM algorithm.}
In the reentrant lines, each queue is served by a single server, so we write $\mu_q$ for
its service rate and $r(q)$ for the queue its jobs move to next. The LLM algorithm gives each
nonempty queue the score
\begin{equation}
S_q(X)=\mu_q\bigl(w_qX_q-\gamma w_{r(q)}X_{r(q)}\bigr)+B_q,
\label{eq:queueing-llm-pressure}
\end{equation}
where $w_q>0$ is a queue weight, $B_q$ is a fixed bonus, and $\gamma$ discounts the downstream
queue. The downstream term is zero when jobs from $q$ leave the network after service. The score
favors serving a queue with a long weighted backlog, but less so when that would feed a downstream
queue that is already long, and the bonus shifts priorities between classes. Each station serves
its highest-scoring nonempty queue, so it never idles while it has work, even if every score is
negative. Given an instance, the \texttt{design} routine sets $w$, $B$ and $\gamma$ by simulating a
small menu of parameter settings and keeping the one with the lowest simulated cost.

\paragraph{Relationship to existing policies.}
The score \eqref{eq:queueing-llm-pressure} is a version of maximum pressure. The basic
maximum-pressure rule scores queue $q$ by $\mu_q(X_q-X_{r(q)})$, the service rate times how much
longer queue $q$ is than the queue it feeds. Each station serves its highest-scoring queue, or idles
if every score is nonpositive. Apart from idling, this is \eqref{eq:queueing-llm-pressure} with
$w_q=1$, $B_q=0$ and $\gamma=1$.

\citet{dai2005pressure} generalize this rule by replacing each queue length $X_q$ with a weighted,
shifted level $w_qX_q-\theta_q$. Expanding their score gives \eqref{eq:queueing-llm-pressure} with
$\gamma=1$ and the constant $\mu_q(\theta_{r(q)}-\theta_q)$ in place of $B_q$. Every route in our
networks ends in an exit, so the offsets $\theta_q$ can be chosen backward from the exit to produce
any bonuses $B_q$. The LLM rule therefore differs from Dai--Lin in only two ways: it discounts the
downstream queue by $\gamma<1$, and it never idles. 

\paragraph{Performance comparison.}
To assess these differences, we compare basic maximum pressure with the
four combinations of $\gamma=1$ or free $\gamma\in[0,1]$ and forced or permitted
idling. We continuously tune positive weights, signed bonuses and, when
free, $\gamma$, extending the original LLM program's finite parameter menu.
\Cref{tab:queueing-mechanism-relative} reports costs relative to basic
maximum pressure. Fitting weights and bonuses reduces costs substantially, while the four fitted variants differ by at most
$0.25\%$ in mean cost and have identical costs on every path for $L=4,\ldots,7$.

\begin{table}[H]
\centering
\small
\setlength{\tabcolsep}{4pt}
\input{tables/results_queueing_mechanism_relative}
\caption{Mean holding cost as a percentage of basic maximum pressure
($100$ in each network; lower is better), from $40$ fresh paired paths per
network. All weighted variants tune both weights and bonuses.
The LLM form uses continuous tuning beyond its original finite menu.
Values are rounded to one decimal; intervals and full results are in
\Cref{app:queueing-mechanism}.}
\label{tab:queueing-mechanism-relative}
\end{table}

The LLM algorithm is a small variant of a policy family already in the
literature: Dai--Lin's weighted maximum-pressure policies. After tuning,
the existing policy achieves essentially the same performance on these
instances. The LLM has therefore produced an effective policy in a slightly
different form, without a material performance advantage over the existing family.

%% file: tables/results_queueing_mechanism_relative.tex
\begin{tabular}{lllrrrrrr}
\toprule
Policy & $\gamma$ & Idling & $L=2$ & $L=3$ & $L=4$ & $L=5$ & $L=6$ & $L=7$ \\
\midrule
Basic & $1$ & Permitted & 100.0 & 100.0 & 100.0 & 100.0 & 100.0 & 100.0 \\
\midrule
Weighted & $1$ & Not allowed & 57.8 & 43.4 & 37.2 & 31.3 & 26.9 & 23.5 \\
Weighted (Dai--Lin) & $1$ & Permitted & 57.8 & 43.4 & 37.2 & 31.3 & 26.9 & 23.5 \\
Weighted (\textbf{LLM form}) & Free & Not allowed & 58.0 & 43.4 & 37.2 & 31.3 & 26.9 & 23.5 \\
Weighted & Free & Permitted & 58.0 & 43.4 & 37.2 & 31.3 & 26.9 & 23.5 \\
\bottomrule
\end{tabular}

%% file: sec_assortment.tex
\section{Assortment Experiments}
\label{sec:assortment}

Our assortment experiments use the hard-instance benchmark of \citet{guo2026benchmark}, whose
instances are constructed to be difficult for the standard heuristics of the assortment
literature. 

\subsection{Instances}
\label{sec:as-instances}

We use three assortment families from the hard-instance benchmark of
\citet{guo2026benchmark}. Each family is treated as its own problem class in the sense of
\Cref{sec:primitives}.

\paragraph{MMNL with cardinality constraint.}
The first family is assortment optimization under the mixed multinomial logit model with a cardinality cap, as in
\Cref{ex:assortment_mmnl}.
We use all
$628$ released instances, which have $n \le 200$ products and $m \le 25$ customer segments.
The level-2 prompt gives the same size bounds: $n \le 200$ products and
$m \le 25$ customer segments.

\paragraph{Nested logit.}
The second family is nested logit. There are $m$ nests of $n$ products each, and a solution is a
binary offer matrix $S$. Nest $i$ has preference
$V_i(S) = v_{i0} + \sum_j S_{ij} v_{ij}$ and is chosen with probability
$V_i(S)^{\gamma_i} / (v_0 + \sum_k V_k(S)^{\gamma_k})$, where $\gamma_i$ is the nest exponent
parameter. The expected revenue is
\[
R(\theta, S) \;=\; \sum_{i=1}^{m} \frac{V_i(S)^{\gamma_i}}{v_0 + \sum_k V_k(S)^{\gamma_k}}
\cdot \frac{\sum_j S_{ij}\, r_{ij} v_{ij}}{V_i(S)} .
\]
We use all $971$ released instances. The released hard set spans
$n \in \{25,50\}$ products per nest and $m \in \{5,10,20\}$ nests, with either no
cardinality constraint or per-nest cap rates in $\{0.1,0.3,0.5\}$.
The level-2 prompt states $n \le 50$ products per nest and $m \le 20$ nests,
with \texttt{cap}, when present, interpreted as a per-nest limit.

\paragraph{MMNL with general linear constraints.}
The third family uses the same MMNL objective but replaces the cardinality cap with five linear
constraints $A x \le B$ on the product indicator vector $x$. The authors provided the generator
and per-method results for their constrained-MMNL experiment. We regenerate their $1{,}800$
instances exactly and drop the six degenerate instances with no feasible product, leaving $1{,}794$
instances for the level-2 comparison. The grid covers every combination of
$m \in \{5,10,25\}$ and $n \in \{50,100,200\}$ under both utility scalings, with 100 seeds per
configuration.
The level-2 prompt states $n \le 200$, $m \le 25$, and five linear constraints.

\subsection{LLM experiments and benchmark algorithms}
\label{sec:as-levels}
\label{sec:as-benchmarks}

Level 1 uses one query per evaluated instance, with a 900-second sandbox budget, so we evaluate it on a smaller stratified set. 
The prompts name the choice model but do not name benchmark methods or suggest a solution approach. 
For level 1, we use a smaller stratified set: the lowest-seed instance from every
released MMNL and nested-logit configuration, giving $72$ MMNL and $48$ nested-logit instances, and
the two lowest seeds from each of the $18$ constrained-MMNL configurations, giving $36$
constrained-MMNL instances.
Additional protocol details are reported in \Cref{app:protocol,app:full-assortment}; representative prompt examples are reported in \Cref{app:prompts}.

\paragraph{Benchmark algorithms.}
We compare the LLM outputs with the benchmark algorithms tested by \citet{guo2026benchmark}, using
per-instance results provided by the authors. Write $S$ for the offered set and $R(S)$ for expected
revenue.
\begin{itemize}[leftmargin=*, itemsep=2pt, topsep=3pt]
\item \emph{Revenue-ordered}: the best prefix of the price ordering. Optimal under plain
multinomial logit \citep{talluri2004}, with no guarantee once segments are mixed. 
For nested logit, we use our own implementation, which jointly optimizes the price-prefix lengths across nests, allowing empty prefixes and respecting each per-nest cap.
\item \emph{Alpha-phi}: the parametric scoring heuristic included in the source benchmark for MMNL
and nested logit.
\item \emph{ADXOpt}: the local-search method of \citet{jagabathula2014}, evaluated by the source
benchmark on all three families.
\item \emph{NN + local search}: the neural-network method of \citet{guo2026benchmark} with its
local-search postprocessing.
\item \emph{Kunnumkal LP}: the LP-based nested-logit policy of
\citet{kunnumkal2023}.
\item \emph{Conic MIP}: the mixed-integer conic formulation reported by
\citet{guo2026benchmark} for MMNL and constrained MMNL.
\end{itemize}
For MMNL, we compare with
revenue-ordered, alpha-phi, ADXOpt, NN + local search, and conic MIP. For nested logit, we compare
with revenue-ordered, alpha-phi, ADXOpt, NN + local search, and Kunnumkal LP. For constrained MMNL,
we compare with revenue-ordered, ADXOpt, NN + local search, and conic MIP.

\subsection{Results}
\label{sec:as-results}

\begin{figure}[!tp]
\centering
\includegraphics[width=\textwidth]{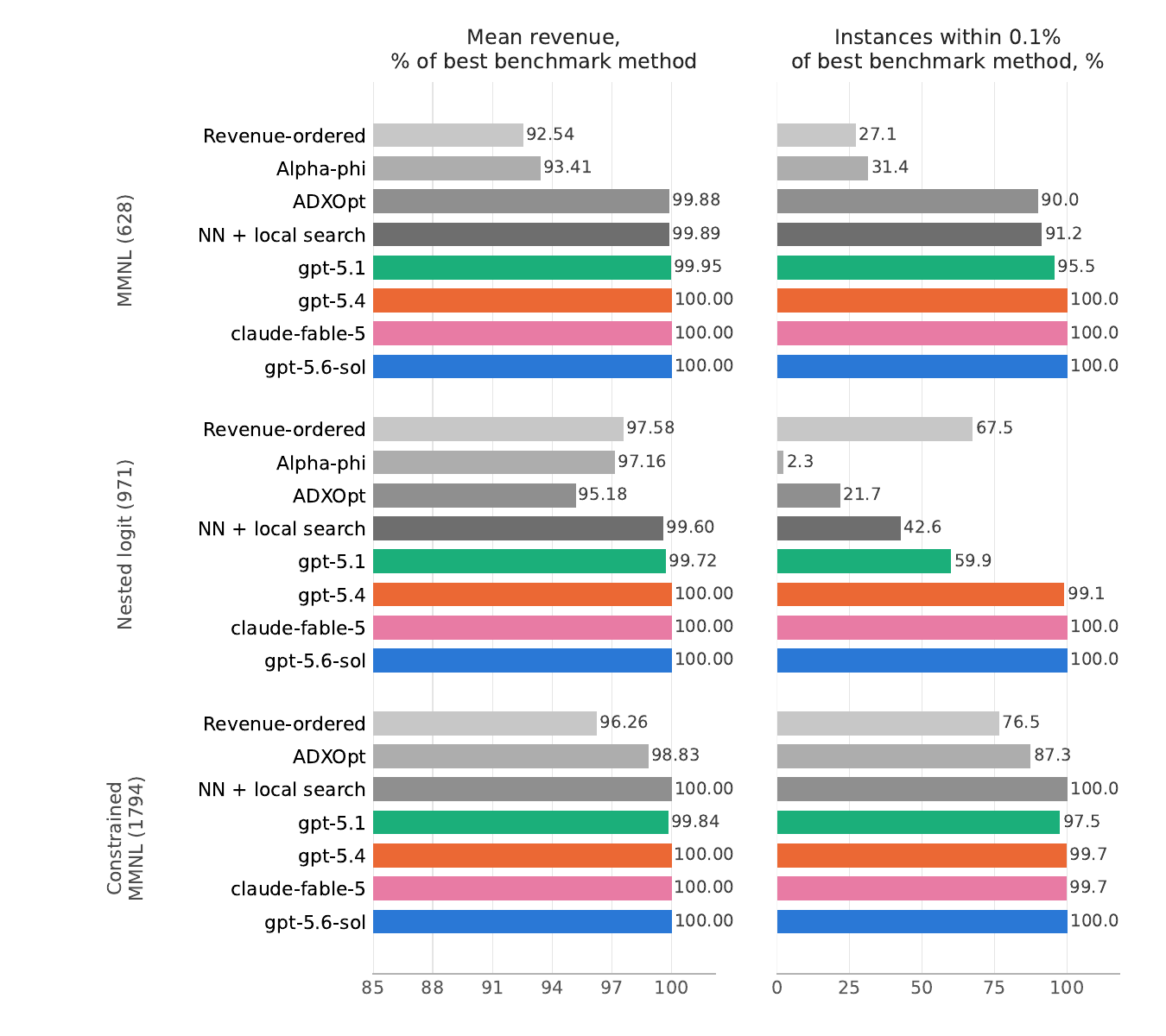}
\caption{Level-2 LLM algorithms against the methods evaluated by \citet{guo2026benchmark} on
identical instances, using the authors' own per-instance results.
Left: mean revenue as a
percentage of the best benchmark method on each instance (axis starts at $85\%$). Right: share of
instances within $0.1\%$ of the best benchmark method. The best benchmark method is the best of
these methods on the instance; it includes the conic solver for MMNL and constrained
MMNL and the LP-based policy of \citet{kunnumkal2023} for nested logit, and never the LLM
solutions. 
}
\label{fig:assortment-vs-published}
\end{figure}

We report each method's revenue as a percentage of the best benchmark method on that instance.
\Cref{fig:assortment-vs-published} gives the level-2 comparison for the three choice models. Each LLM is queried once
per problem family and returns a solver, which we then run on every instance in that family. The
methods of \citet{guo2026benchmark} are evaluated on the same instances using the authors' per-instance results.

Across all three families, the strongest level-2 LLM solvers
essentially match the best benchmark method. On MMNL, \model{gpt-5.6-sol},
\model{claude-fable-5}, and \model{gpt-5.4} attain the benchmark optima on
all $628$ instances; the next subsection explains the special structure
that makes these instances tractable. On constrained MMNL,
\model{gpt-5.6-sol} matches the best benchmark method on every instance,
while \model{claude-fable-5} and \model{gpt-5.4} fall outside the
$0.1\%$ tolerance on only a few instances. The simpler classical
heuristics have substantially larger losses on the tail. 

On nested
logit, \model{gpt-5.6-sol} and \model{claude-fable-5} match the best
existing method within $0.1\%$ on all $971$ instances, with mean revenue
$100.00\%$ of the comparator.
The nested-logit comparator is already very close to optimal. It
includes the LP-based policy of \citet{kunnumkal2023}, whose revenue
is within $0.1\%$ of their upper bound on $911$ of the $971$ instances
and within $0.5\%$ on every instance. 

\Cref{fig:assortment-vs-published-l1} in
\Cref{app:full-assortment} gives the level-1 counterpart on the smaller stratified set defined in
\Cref{sec:as-levels}.
On these subsets, \model{gpt-5.6-sol} and \model{claude-fable-5}
attain the same revenues at levels 1 and 2 on every evaluated instance. For these models, giving the full instance
and a separate query per instance yields no improvement over the
reusable level-2 solver.
The main visible benefit of level 1 is for
\model{gpt-5.1} on nested logit, where instance-specific querying raises the within-$0.1\%$ share
from $60.4\%$ to $87.5\%$ on the same subset.

\subsection{The two-type MMNL structure}
\label{sec:as-mmnl-structure}

The MMNL results should be read with an important caveat. All $628$ released MMNL instances have
only two distinct product-utility vectors. Thus, if this structure is recognized, it is easy to
find the exact optimal assortment: sort each utility class by price, and enumerate how many
products to take from each class. Our independent solver uses this enumeration and confirms the
benchmark optima on all $628$ instances.

It is still interesting that level-2 algorithms solve this benchmark, because the prompt gives no
example instances and does not mention the two-type structure. Among the published heuristic and
learning baselines, no method exploits this collapse well enough to match the optimum on every
instance; the LLM solvers do. At the same time, the result is less strong than it first appears:
the algorithms are not solving general MMNL in full generality; they are solving a highly
structured two-type benchmark.

\subsection{Structure of the LLM algorithms}
\label{sec:as-sol-l2}

We inspect the selected \model{gpt-5.6-sol} level-2 algorithms to understand what they are doing. In assortment optimization the returned object is not a policy, as in inventory, but a solver: executable code that maps a new instance to one assortment. We describe only the selected \model{gpt-5.6-sol} solvers here; \Cref{app:reasoning} gives more detail on the model's reasoning.

\paragraph{MMNL with cardinality constraint.}
The \model{gpt-5.6-sol} MMNL solver is a heuristic portfolio with exact branches for easy cases. It solves one-segment and small-enumeration cases directly; otherwise it generates candidate assortments from score-ordered prefixes, greedy add and delete paths, randomized starts, and a continuous relaxation. It then polishes the best candidates with add, drop, and swap local search. Thus the solver is stronger than a single revenue-ordered or greedy rule, but it is not a general exact MMNL algorithm; its exactness on the released benchmark should be read together with the two-type structure in \Cref{sec:as-mmnl-structure}.

\paragraph{Nested logit.}
The \model{gpt-5.6-sol} nested-logit solver uses the separable structure of the model. It summarizes each nest-level subset by its numerator and attraction contribution, builds a pool of candidate subsets for each nest, and then chooses one subset per nest using a fractional-programming update. For a current revenue target, the cross-nest step scores each candidate by ``numerator minus target times attraction,'' then updates the target from the resulting full assortment. This uses ideas from nested-logit assortment algorithms and bounds \citep{davis2014nested,kunnumkal2023}, but with a richer generated menu than price-ordered prefixes alone.

\paragraph{MMNL with general linear constraints.}
The \model{gpt-5.6-sol} constrained-MMNL solver is a resource-aware portfolio. It filters individually infeasible products, repairs candidates to satisfy the linear constraints, generates greedy and relaxation-based starts, and improves feasible candidates by add, drop, and swap moves. When time permits, it also solves continuous relaxations and restricted MILP subproblems around the incumbent. The code therefore combines standard OR primitives: relax, round, repair, locally improve, and occasionally solve a small exact subproblem.

%% file: sec_stability.tex
\section{Robustness and Sensitivity Checks}
\label{sec:robustness}

The main results use a fixed compute budget and a single LLM query per result.
This section checks whether the results are sensitive to query-time compute
(\Cref{sec:query-time-compute}), whether they extend to new instances
(\Cref{sec:l2-holdout}), and how much they vary across repeated runs of the same LLM query
(\Cref{sec:l2-stability}).

\subsection{The role of query-time tools}
\label{sec:zero-compute}
\label{sec:query-time-compute}

We remove sandbox compute entirely and measure how much the results change. We rerun selected
\model{gpt-5.6-sol} queries with the compute budget set to zero. The prompt is unchanged except
that it states a budget of 0 seconds, so the model cannot execute Python to run simulations or test a policy; it must write its answer from reasoning alone.

The main result is a contrast between levels. At level 2, zero query-time compute works surprisingly
well for inventory and assortment: the model can write strong reusable algorithms without running
simulations or testing candidates during the query.
For queueing, level-2 zero-compute remains strong on the criss-cross class, but performs worse on the N-model and reentrant-line classes.
At level 1, the same removal is much more damaging in the inventory experiment, where most zero-compute policies fall back to simple closed-form heuristics and perform worse than the tuned benchmark.

\paragraph{Inventory.}
\Cref{tab:zero-compute} reruns the deterministic lost-sales lead-time sweep with zero query-time
compute. At level 2, the zero-compute algorithm beats the best tuned benchmark on all six
lead-time instances and is within $0.4$ percentage points of the full-budget algorithm on every
one of them. At level 1, removing the Python tool hurts much more: five of the six zero-compute
policies are simple closed-form heuristics and trail the benchmark by $3.5\%$ to $13\%$.

\begin{table}[t]
\centering
\input{tables/results_zero_compute}
\caption{Lead-time sweep with the sandbox compute budget set to zero (\model{gpt-5.6-sol}). Entries are the percentage gap versus the best tuned benchmark per instance (negative = better, in \textbf{bold}); the 3600\,s rows repeat the main results for comparison.}
\label{tab:zero-compute}
\end{table}

\paragraph{Assortment.}
We also rerun the level-2 assortment queries with zero query-time compute.
\Cref{tab:assortment-zero-compute} reports the results using the same comparator as
\Cref{fig:assortment-vs-published}: mean revenue as a percentage of the best benchmark method on
each instance, and the share of instances on which the algorithm is no worse than it. For all three
assortment families, removing sandbox compute leaves mean performance unchanged at the precision
we report.

\begin{table}[t]
    \centering
    \input{tables/results_assortment_zero_compute}
    \caption{Zero-compute level-2 assortment results for \model{gpt-5.6-sol}. The prompt gives no query-time Python budget. Entries are mean revenue as a percentage of the best benchmark method on each instance, the comparator of \Cref{fig:assortment-vs-published}, and the share of instances on which the algorithm is within $0.1\%$ of or better than it, computed on the same instances in both columns.}
    \label{tab:assortment-zero-compute}
    \end{table}

\paragraph{Queueing.}
The queueing zero-compute results are more mixed.
\Cref{tab:q-zero-compute} reports the mean per-instance percentage cost gaps.
For the criss-cross class, the zero-compute
artifact is essentially optimal because its returned \texttt{design} routine can solve the small
dynamic program at evaluation time. The N-model artifact is $9.74\%$ above optimum, compared with $0.003\%$ with compute. On the reentrant-line instances, zero compute remains
usable but is about $2.4\%$ above the PPO comparator on average, while the full-budget level-2
algorithm beats PPO by about $5.0\%$. This suggests that zero query-time compute can work when the
returned code can still do the relevant optimization later, but query-time simulation helps on the
larger or less exact queueing classes.

\begin{table}[t]
\centering
\input{tables/results_queueing_zero_compute}
\caption{Full-budget and zero-compute level-2 queueing results for \model{gpt-5.6-sol}. Entries are mean per-instance percentage cost gaps relative to the exact optimum (criss-cross and N-model) or PPO (reentrant line); negative is better. Bold marks a mean gap of at most $0.1\%$.}
\label{tab:q-zero-compute}
\end{table}

\paragraph{Adding a web-search tool.}
The zero-compute check removes a resource from the query; we also tried adding one. We reran four level-2 queries (deterministic lost sales, MMNL, nested logit, and constrained MMNL) for \model{gpt-5.6-sol} with an unlimited web-search tool attached alongside the sandbox, all else unchanged. The model used the tool only as a brief literature check at the start of the session, or not at all, and the results match the main artifacts on all four classes. On nested logit, the algorithms with and without web search both match the best benchmark method within $0.1\%$ on every instance. \Cref{app:websearch} reports the searches and results.

\subsection{Generalization to new instances}
\label{sec:l2-holdout}

For inventory and assortment, we ask whether the level-2 algorithms generalize beyond the
evaluation grids used in the main experiments. This check helps address benchmark contamination:
if the main results only reflected memorization of the public benchmark grids, the same frozen code
would have less reason to work on these new instances. We use the level-2 algorithms from the main
runs and evaluate the same returned code on the new instances; no model is queried again.

For lost-sales inventory, the new instances use different demand means, penalty costs, and lead-time ranges; benchmarks are
tuned separately for each new instance. For assortment, we generate continuous-utility MMNL
instances to remove the two-utility-column structure in the published MMNL benchmark, and large
synthetic nested-logit instances that avoid the revenue-ordered-friendly regime.

\begin{figure}[t]
\centering
\includegraphics[width=0.99\textwidth]{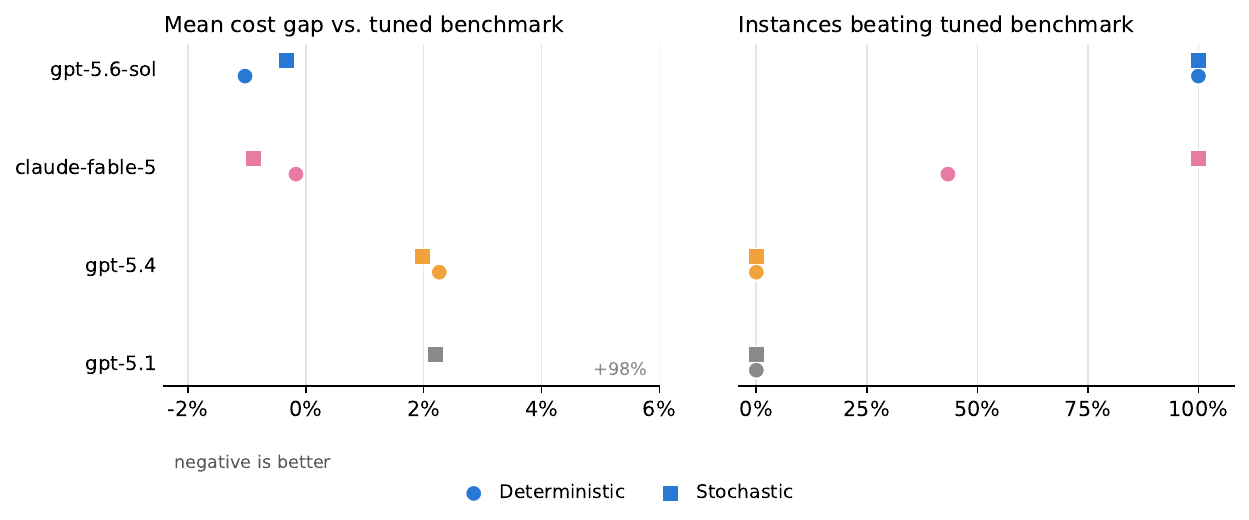}
\caption{Inventory holdout results for frozen level-2 artifacts. The left panel reports the
mean cost gap versus the best tuned benchmark on new lost-sales parameter values; smaller values
are better. The right panel reports the share of holdout instances on which the level-2 algorithm
beats that benchmark; larger values are better. The deterministic \model{gpt-5.1} point is outside the plotted range at
$+98.4\%$.}
\label{fig:l2-holdout-inventory}
\end{figure}

\Cref{fig:l2-holdout-inventory} shows the main inventory pattern. \sol continues to beat the
best tuned benchmark on every deterministic and stochastic holdout instance. \model{claude-fable-5}
is also strong: it beats every stochastic holdout and 13 of the 30 deterministic holdouts. The older models
do not show the same robustness. \model{gpt-5.4} is modestly worse than the best benchmark on most
holdout cells, while \model{gpt-5.1} fails badly on deterministic holdouts.

The assortment holdouts show the same broad pattern. 
For MMNL, the level-2 algorithms from all four models match the comparator revenue on every generated instance, which supports that they are not merely exploiting the two-type structure of the published benchmark.
For nested logit, all four artifacts also match the multistart local-search comparator throughout the holdout. 
These results support generalization, but they are not definitive stress tests: the generated
MMNL instances are still easy for a greedy-local heuristic, and the nested-logit comparator is
heuristic rather than certified. Detailed holdout construction and aggregate assortment
results are in \Cref{app:l2-holdout-results}.

\subsection{Sensitivity to repeated draws}
\label{sec:l2-stability}

The main results report one LLM query for each model and problem class. To check robustness across identical queries, we run additional \sol level-2 queries for each inventory, structurally specified queueing, and assortment class.
Each draw used the same problem-class prompt and the same evaluation protocol. We call an algorithm defective if it fails to produce valid results under our evaluator.

\begin{figure}[H]
\centering
\includegraphics[width=1\textwidth]{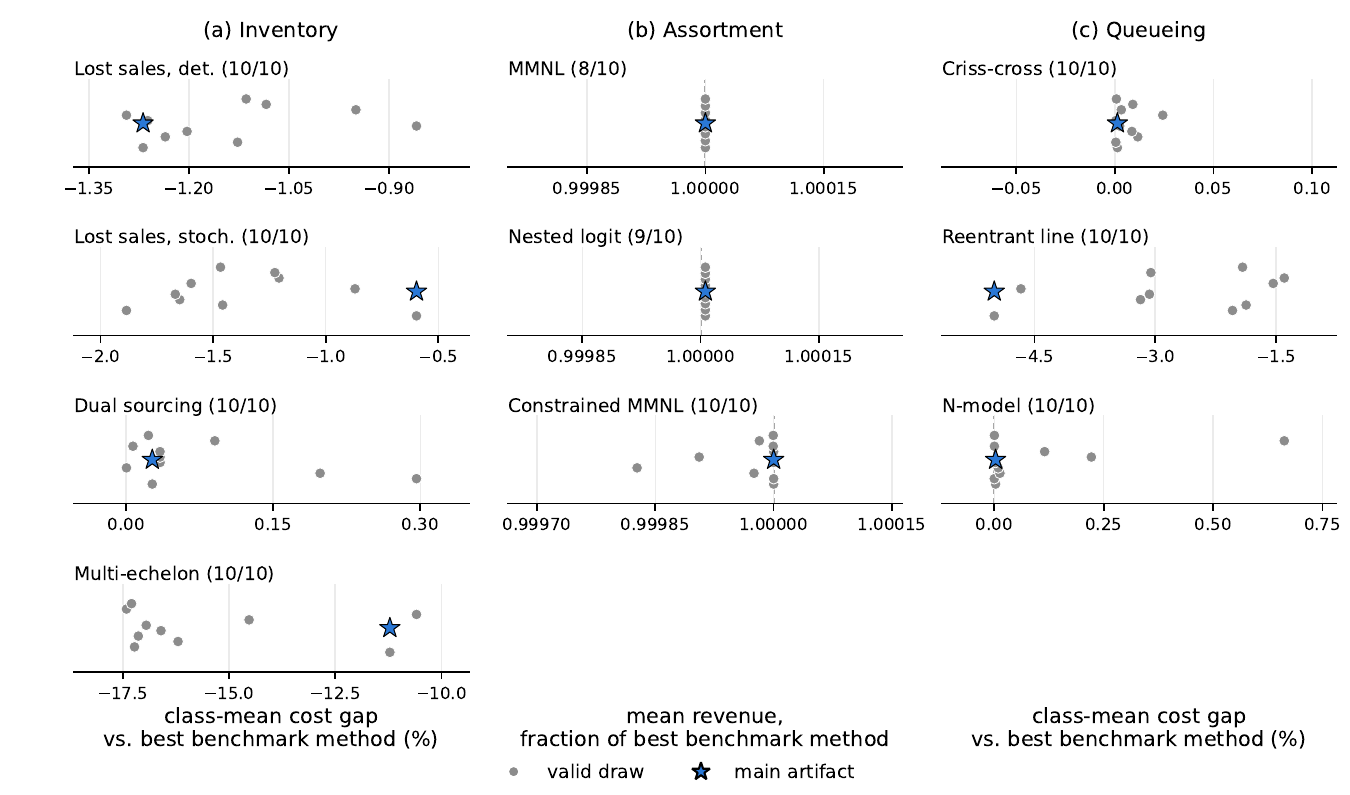}
\caption{Repeated level-2 draws for \sol on inventory, assortment, and queueing classes.
Each gray dot is one valid algorithm from
the ten-draw stability study; the blue star marks the artifact used in the main results.
Each row
uses its own x-axis scale to show within-class variation. Inventory reports class-mean cost gap to
the best benchmark method (the best tuned benchmark, or the exact optimum on dual sourcing).
Assortment reports mean revenue as a fraction of the best benchmark method, the comparator of
\Cref{fig:assortment-vs-published}. Queueing reports class-mean cost gap to the best benchmark method,
computed by exact policy evaluation against the optimum for the criss-cross and N-model classes
and by simulation against the per-instance PPO policy for the reentrant-line class. Defective
algorithms are counted in the panel labels but not plotted.}
\label{fig:l2-stability-dots}
\end{figure}

\Cref{fig:l2-stability-dots} plots the repeated level-2 draws, with the artifact used in the main results marked separately. The main pattern is that the selected artifacts are not outliers. Across inventory and assortment, most valid draws fall in a narrow performance range: the class-mean spread is below $1.3$ percentage points for both lost-sales classes, the nine valid nested-logit draws are identical at $100.00\%$ mean revenue, and constrained MMNL varies by at most $0.02$ percentage points in mean revenue. The two main qualifications are multi-echelon inventory, where valid draws differ by $6.8$ percentage points, and dual sourcing, where
seven of ten draws have a mean absolute optimality gap below $0.05\%$, while the other three have mean gaps below $0.3\%$.

The queueing repeats show a similar pattern for the structurally specified level-2 classes. 
Every criss-cross draw has a mean optimality gap below $0.03\%$ across the six instances.
    For the reentrant-line class, all ten draws beat PPO on average over the six instances, with mean gaps between $-1.4\%$ and $-5.0\%$. 
All ten N-model draws have converged optimality gaps below $0.7\%$; all ten artifacts attempt dynamic programming, but differ in their implementation and stopping rules.

Overall, the stability study suggests that the main results are not driven by a single lucky draw. The main failure mode is not large performance variation among valid algorithms, but occasional defective or non-evaluable artifacts: three of the 100 pooled draws remain defective, all in assortment.

%% file: tables/results_zero_compute.tex
\begin{tabular}{llrrrrrr}
\toprule
\textbf{Level} & \textbf{Compute} & $\ell=5$ & $\ell=6$ & $\ell=7$ & $\ell=8$ & $\ell=9$ & $\ell=10$ \\
\midrule
\multirow{2}{*}{Level 1} & 3600 s & \textbf{-2.50} & \textbf{-3.23} & \textbf{-4.06} & \textbf{-3.46} & \textbf{-3.11} & \textbf{-2.78} \\
\cmidrule(l){2-8}
 & 0 s & +3.51 & \textbf{-3.07} & +8.80 & +10.98 & +12.97 & +6.32 \\
\midrule
\multirow{2}{*}{Level 2} & 3600 s & \textbf{-2.45} & \textbf{-3.08} & \textbf{-4.06} & \textbf{-3.56} & \textbf{-3.08} & \textbf{-2.73} \\
\cmidrule(l){2-8}
 & 0 s & \textbf{-2.40} & \textbf{-3.14} & \textbf{-3.75} & \textbf{-3.23} & \textbf{-2.94} & \textbf{-2.69} \\
\bottomrule\end{tabular}

%% file: tables/results_assortment_zero_compute.tex
\small
\setlength{\tabcolsep}{6pt}
\begin{tabular}{@{}lcccc@{}}
\toprule
Family (\# instances) & \multicolumn{2}{c}{Mean revenue, \% of best benchmark} & \multicolumn{2}{c}{No worse than best benchmark, \%} \\
\cmidrule(lr){2-3}\cmidrule(lr){4-5}
 & With compute & No compute & With compute & No compute \\
\midrule
MMNL (628) & $100.00\%$ & $100.00\%$ & $100.0\%$ & $100.0\%$ \\
Nested logit (971) & $100.00\%$ & $100.00\%$ & $100.0\%$ & $100.0\%$ \\
Constrained MMNL (1,794) & $100.00\%$ & $100.00\%$ & $100.0\%$ & $99.9\%$ \\
\bottomrule
\end{tabular}

%% file: tables/results_queueing_zero_compute.tex
\begin{tabular}{lrr}
\toprule
\textbf{Class} & \textbf{Full L2} & \textbf{Zero L2} \\
\midrule
Criss-cross & $\mathbf{0.00\%}$ & $+0.13\%$ \\
Reentrant line & $\mathbf{-5.00\%}$ & $+2.39\%$ \\
N-model & $\mathbf{0.00\%}$ & $+9.74\%$ \\
\bottomrule
\end{tabular}

%% file: sec_conclusion.tex
\section{Conclusion}
\label{sec:conclusion}

This paper asks whether a general-purpose LLM can design algorithms for well-specified
operations problems. Across inventory control, queueing network control, and assortment
optimization, the strongest model matches or outperforms the best benchmark method on nearly all
evaluated instances. The level-2 result matters most: a single query produces an algorithm that
works across a whole problem class, and for inventory and assortment the same code keeps
performing well on new instances outside the original benchmarks (\Cref{sec:l2-holdout}).
Performance also improved sharply across models released within eight months of each other.

The mechanism case studies clarify the nature of these results. For
deterministic lost-sales inventory, the LLM extends projected inventory-level
control with a tuned standard-deviation term, yielding a small but consistent
improvement. For reentrant queues, its policy is a small variant of weighted
maximum pressure, and the existing policy family achieves essentially the
same performance after tuning. These cases show useful adaptations of
existing ideas, but do not demonstrate major new algorithmic principles.
Identifying a suitable policy family, implementing it, and tuning it are
nevertheless valuable tasks. 
LLMs can help translate established algorithmic ideas into effective solutions for particular problems.

Real operations problems still require formulation, measurement, validation, deployment, and judgment about objectives and trade-offs. 
This paper isolates only the algorithm-design step. 
The results show that LLMs can produce effective algorithms for the problem classes we study, but leave open how well they can develop new algorithmic principles.
Understanding when this succeeds, when it fails, and how to validate the resulting algorithms is an important direction for future work.

%% file: sec_appendix.tex
\section{Protocol details}
\label{app:protocol}

\paragraph{Models and hardware.}
We use four models: \model{gpt-5.6-sol}, \model{gpt-5.4}, and \model{gpt-5.1} from
OpenAI, and \model{claude-fable-5} from Anthropic. The date-stamped served
identifiers in our logs are \model{gpt-5.1-2025-11-13} and
\model{gpt-5.4-2026-03-05}. The logged identifiers for \model{gpt-5.6-sol} and
\model{claude-fable-5} did not include date suffixes, so we report them by alias.
All local computation, including sandbox executions, benchmark tuning, simulation scoring, and returned-code evaluation, ran on one laptop: an Apple M2 Pro with 16\,GB of memory.

\paragraph{Budgets.}
Each query has a sandbox compute budget, and after every code execution the model is
told how much remains. The budget is 3600 seconds per query, except for level-1 assortment
queries, which receive 900 seconds because evaluation is closed-form and pilot runs used far less.
We report used compute for every query. Returned level-2 algorithms have a stated 30-second
per-instance setup target on one CPU core; level-1 policies may do analogous setup at module load
under the same target. Inventory prompts also state that returned \texttt{order} functions should
be lightweight. The 30-second rule is not enforced, but we measure and report these times.

\paragraph{Level-specific prompt contracts.}
In inventory, level 1 gives the numeric instance parameters and asks for Python code defining
\texttt{order(on\_hand, pipeline)} for that instance. Level 2 gives the problem class and parameter
ranges, but no example instances or evaluation grid, and asks for \texttt{design}, which maps
instance parameters to an \texttt{order} function. For lost-sales classes, the provided ranges are
$\ell \in [2,12]$, $\lambda \le 50$, and $p \le 100$ for deterministic lead times, and
$\ell_{\min}\ge 2$, $\bar\ell \le 12$, $\lambda \le 50$, and $p \le 100$ for stochastic lead
times.
In queueing, level 1 gives the numeric network parameters and asks for
\texttt{policy(queues, t)}. Level 2 gives a subclass description and parameter ranges, but no
evaluation instances, and asks for \texttt{design(network, mu, lam, next\_queue, h)}, which returns
a scheduling policy for a future instance.
In assortment, level 1 gives the full numeric instance arrays and asks for one
assortment. Level 2 gives the choice-model family and broad size ranges, but no benchmark instance,
and asks for Python code defining \texttt{solve}. Each returned level-2 solver is evaluated on every
instance in its family. The 30-second runtime target is stated in the prompt and measured in
evaluation; a 600-second guard catches solvers that never return.

\paragraph{Scoring and reporting.}
  Every comparison is to the best benchmark method on the instance, as defined in
\Cref{sec:protocol}; LLM solutions never enter the comparator. Inventory policies are simulated on common random numbers and scored on streams separate from tuning. Reported confidence intervals describe individual policy costs. 
For assortment, we use the authors' per-instance revenues for their methods and our evaluator for
the LLM solvers. For queueing, we evaluate policies exactly on the DP-solvable instances
and by steady-state simulation on the reentrant-line instances.
Logs include the prompt, model reasoning and code, execution outputs
and durations, token usage, and served model version. If a model artifact is defective, we rerun
the same prompt to obtain an evaluable artifact, preserve the defective artifact, and report the
defect where it affects the results.
All logs are released together with the code and prompts at \url{https://anonymous.4open.science/r/llm-or-algorithms-F9F2}.

\section{Additional inventory details}
\label{app:validation}
\label{app:additional-inventory}

This appendix gives the inventory validation checks, additional inventory settings, full
per-instance lost-sales results, inventory timing results, and the projected-inventory
mechanism experiment.

\subsection{Validation}

Before scoring any inventory policy, we validate the relevant simulator and benchmark
implementation.

\paragraph{Benchmark tuning.}
For each benchmark policy with free parameters, every candidate is evaluated by simulation and the
best candidate is kept. Search ranges are written relative to the instance parameters, with
order-up-to levels around mean lead-time demand and caps and constant order rates around mean
demand, so one specification serves every instance. For lost sales, base-stock levels and caps are searched at integer resolution. For each cap, we evaluate a coarse grid of stock levels and then every integer stock level between the neighbors of its best coarse point. Mixed strategies search both $(r-1,r)$ and $(r,r+1)$ around the best stable integer constant order $r$, with mixing probabilities spaced by $0.05$. Constant and mixed orders are restricted to mean order rates strictly below mean demand. The same search implementation is used for the main and holdout grids, and tuning and final scoring use disjoint random-number streams.

\paragraph{Lost sales.}
The simulator retains all arriving inventory and limits only the quantity ordered each period.
On the short lead-time instances, where the exact optimum is computable, simulated long-run costs
match the dynamic program within simulation error (differences of at most $0.009$ on costs of
order five). The stochastic lead-time simulator is validated by checking the degenerate
deterministic case, Little's law, and the realized lead-time distribution. Where the source paper's
text and figure disagree on the stochastic lead-time range ($\bar\ell$ starting at 4 or 5), we
follow the figure. Our tuned benchmarks also reproduce the qualitative pattern in the source
paper's Figure~5.

\paragraph{Dual sourcing.}
Our tuned benchmarks reproduce the source paper's Figure~9 to within a
hundredth of a percentage point on all but two of its eighteen policy-instance pairs.

\paragraph{Multi-echelon.}
In this setting, there are no published numbers to reproduce. \citet{vanroy1997} do not specify their simulation in
enough detail, and \citet{gijsbrechts2022drl} report only improvements relative to their own tuned
base-stock policy. The simulator is therefore validated internally: a scalar, line-by-line
transcription of the source paper's equations matches the vectorized simulator exactly on shared
random streams, and in a limiting parameter regime (one retailer, no special delivery, a warehouse
that always fills orders) the system reduces to single-location lost sales, whose exact long-run
cost the simulator matches to within simulation error.

\subsection{Dual sourcing}
\label{app:dual-sourcing}

\paragraph{Instances and dynamics.}
The dual-sourcing setting follows Section~6 of \citet{gijsbrechts2022drl}. A firm stocks one
product and can order from a regular supplier and an expedited supplier. The regular supplier is
slower and cheaper, with unit cost $100$ and lead time $\ell_r \in \{2,3,4\}$. The expedited
supplier arrives immediately and has unit cost $c_e \in \{105,110\}$. Demand is uniform on
$\{0,\ldots,4\}$, unmet demand is backlogged, the holding cost is $5$, and the backlog cost is
$495$, giving six instances. For all six, an exact dynamic program is computationally feasible
and is reported by the source paper, so it is the best benchmark method on these instances and all gaps are relative to the exact optimum.

\paragraph{Benchmarks.}
We compare against three validated dual-sourcing heuristics from the source paper's Figure~9. The
\emph{tailored base-surge} policy \citep{allon2010,xin2018tbs} places a constant regular order and uses the expedited supplier
to raise the short inventory position to a base-stock level. The \emph{dual-index} policy \citep{veeraraghavan2008} uses two
base-stock levels, one for the short position and one for the long position. The \emph{capped
dual-index} policy \citep{sun2019cdi} adds a cap on the regular order. We exclude the single-index policy from the
reported benchmark suite because our reconstruction does not reproduce the source paper's Figure~9
and it is never the binding comparator.

\paragraph{Results.}
Table~\ref{tab:full-dual-sourcing} reports the exact optimum for each of the six instances, and every LLM policy's gap above it at both levels.
\begin{table}[p]
\centering
\input{tables/results_dual_sourcing_llm}
\caption{Dual sourcing LLM policies. Entries are percentage cost gaps relative to the optimal policy evaluated on the same scoring streams (first column); \textbf{bold} marks a gap within $0.1\%$. All 95\% CI half-widths are below 0.22.}
\label{tab:full-dual-sourcing}
\end{table}

\subsection{Multi-echelon distribution}
\label{app:multi-echelon}

\paragraph{Instances and dynamics.}
The multi-echelon setting follows Section~7 and Table~4 of \citet{gijsbrechts2022drl}, adapted from
\citet{vanroy1997}. One warehouse supplies ten identical stores. The warehouse replenishes from a
manufacturer, each store replenishes from the warehouse, and store customers who face a stockout may
accept same-day special delivery from warehouse inventory with probability $0.8$. Otherwise the sale
is lost. Holding costs are $3$ at both the warehouse and each store, the lost-sale penalty is $60$,
production capacity is $100$, the warehouse inventory-position cap is $1000$, and each store's
inventory-position cap is $100$.

We use the two parameter settings printed in Table~4 of \citet{gijsbrechts2022drl}:
\[
    (\ell_w,\ell_r,\mu,\sigma) \in \{(2,2,5,14),\ (5,3,0,20)\}.
\]
Demand at each store is an independent rounded normal random variable, truncated below at zero. In
the second setting, the printed value $\mu=0$ is kept as written; after truncation, the effective
demand mean remains positive. Exact optimization is intractable in this system.

\paragraph{Benchmarks.}
The benchmark is the tuned constant order-up-to policy used by \citet{gijsbrechts2022drl}, with two
parameters: a warehouse order-up-to level and a store order-up-to level, shared across the ten
identical stores. We tune both parameters by simulation on random-number streams separate from the
final scoring streams.

\paragraph{Results.}
Table~\ref{tab:full-multi-echelon} reports the tuned benchmark's cost on both settings and every
LLM policy's gap versus it at both levels.
\begin{table}[p]
\centering
\input{tables/results_multi_echelon}
\caption{Multi-echelon distribution. LLM entries are percentage gaps versus the tuned constant order-up-to benchmark (negative = better, in \textbf{bold}). All 95\% CI half-widths are below 2.6.}
\label{tab:full-multi-echelon}
\end{table}

\subsection{Per-instance inventory results}
\label{app:full}

The six exactly solvable short-lead-time lost-sales instances are reported in
\Cref{tab:full-tier-a} in the main text. \Cref{tab:full-lost-sales} reports the remaining 20
lost-sales instances: four
models at level 1, and each model's two level-2 algorithms evaluated on every instance of its
class. Level-1 and level-2 entries in the same row are directly comparable: all policies in a row
are simulated on identical demand (and lead-time) streams, against the same tuned benchmarks.
The dual-sourcing and multi-echelon tables are in \Cref{app:additional-inventory}.

All entries are percentage gaps versus the best benchmark method on the row, negative
meaning the LLM policy has lower cost. On the short lead-time grid (\Cref{tab:full-tier-a}) and
dual sourcing (\Cref{tab:full-dual-sourcing}) that method is the exact dynamic program, so entries
are gaps above the optimum and \textbf{bold} marks a simulated cost gap within $0.1\%$ of that comparator; in \Cref{tab:full-lost-sales} and multi-echelon (\Cref{tab:full-multi-echelon}) it is the best tuned benchmark on that row and \textbf{bold} marks a cost reduction exceeding $0.1\%$.

\begin{table}[p]
\centering
\input{tables/results_lost_sales}
\caption{Lost sales: per-instance results for the 20 instances beyond the exactly solvable short lead-time grid (\Cref{tab:full-tier-a} in the main text covers that grid). Benchmark columns are long-run average costs, best per row underlined; the mixed strategy is defined for deterministic lead times only. LLM entries, level~1 (L1) and level~2 (L2) per model, are the percentage gap versus the best benchmark on that row (negative = better, in \textbf{bold}). All LLM cost estimates have 95\% CI half-widths below 0.09.}
\label{tab:full-lost-sales}
\end{table}

\subsection{Compute and wall time}
\label{app:inventory-timing}

\Cref{tab:inventory-timing-l2,tab:inventory-timing-l1} report three clocks for
\model{gpt-5.6-sol}. \emph{Wall} is the time from sending the prompt to the final answer,
including reasoning and sandbox execution. \emph{Sandbox} is the metered Python execution time
used of the budget. The third clock measures returned-code setup at evaluation: \texttt{design}$(\theta)$
per instance at level 2, and module-load time at level 1, both against the $30$ s target stated in
the prompt. \Cref{tab:zero-compute-usage} gives the per-instance breakdown for the lead-time sweep
of \Cref{sec:zero-compute}.

Level 1 uses most of its $3{,}600$ s sandbox budget on lost sales and multi-echelon
(means of $1{,}959$ to $2{,}841$ s, with several queries at the cap), where a policy has to be
tuned by simulation. It uses almost none on dual sourcing, where the model recognizes a small exact
dynamic program. Level 2 uses a small fraction of the sandbox budget ($18$ to $237$ s of
$3{,}600$) because it defers computation to \texttt{design}$(\theta)$, which then runs $3$ to
$13$ s per instance at evaluation. The zero-compute queries take three to five minutes of
reasoning and return artifacts whose evaluation-time work is the same size.

\begin{table}[H]
\centering
\small
\input{tables/results_inventory_timing_l2}
\caption{Level-2 inventory queries and returned algorithms, \model{gpt-5.6-sol}: one class-level
query per row at the full budget, and the zero-compute lost-sales query of \Cref{sec:zero-compute}.
Output tokens include reasoning tokens. \texttt{design} runtime is per evaluation instance (mean and
maximum over the class).}
\label{tab:inventory-timing-l2}
\end{table}

\begin{table}[H]
\centering
\small
\input{tables/results_inventory_timing_l1}
\caption{Level-1 inventory queries, \model{gpt-5.6-sol}: distribution over the instances of each
class (one query per instance) of query wall time, sandbox seconds used of the $3{,}600$ s budget,
output tokens including reasoning tokens, and the module-load time of the returned policy at
evaluation. The last row is the zero-compute rerun of the lead-time sweep.}
\label{tab:inventory-timing-l1}
\end{table}

\begin{table}[H]
\centering
\input{tables/results_zero_compute_usage}
\caption{Resource use behind \Cref{tab:zero-compute}, in seconds. Sandbox compute is the metered execution time the model chose to use, and wall clock the full generation time of the query including reasoning; at level 2 these are properties of the single class-level query, so one value covers all six instances. Module load is the setup time when a returned level-1 policy is first imported at evaluation; \texttt{design}$(\theta)$ is the level-2 analog, the per-instance time to construct the policy at evaluation.}
\label{tab:zero-compute-usage}
\end{table}

\input{app_inventory_mechanism}

\section{Additional queueing details}
\label{app:queueing}

This appendix gives the queueing instance definitions, validation checks, timing, and robustness
results behind \Cref{sec:queueing}.

\subsection{Instance definitions}
\label{app:q-instances}

The queueing instances are the thirteen multiclass queueing networks of
\citet{dai2022queueing}. A network has $Q$ queues and $S$ servers. Server $s$ can serve queue $q$
only if the compatibility matrix has $A_{sq}=1$, and then works at rate $\mu_{sq}$. External
arrivals to queue $q$ are Poisson with rate $\lambda_q$. After service completion at queue $q$, a
job either leaves the system or joins the downstream queue \texttt{next\_queue}$(q)$ with a fresh
exponential service requirement. Holding costs are linear in queue lengths.

The six criss-cross instances have two servers, three queues, common arrival rate
$\lambda \in \{0.3,0.6,0.9\}$, and service-rate vector $(2,\mu_2,2)$, with $\mu_2=1$ in the
balanced variants and $\mu_2=1.5$ in the imbalanced variants. The N-model has two servers and two
queues, arrival rates $(1.235,0.38)$, service rates $1$ for server 1 on class 1 and $(0.5,1)$ for
server 2 on classes 1 and 2, holding costs $(3,1)$, and load $0.95$. These seven small instances
have exact dynamic programs. The six reentrant-line instances (the extended six-class networks of \citet{dai2022queueing}) have
$L=2,\ldots,7$ stations and $3L$ queues; they are too large for exact dynamic programming, so the
primary comparator is the per-instance PPO policy of \citet{dai2022queueing}.

\subsection{Simulation and comparator validation}
\label{app:q-validation}

  On the seven DP-solvable instances (criss-cross and N-model), the reported costs use finite-chain evaluation without simulation. For criss-cross, relative value iteration uses queue-length caps $N=60$ in light and medium traffic and $N=130$ in heavy traffic. The computed optima match the published values to three or four digits except in the truncation-sensitive balanced-heavy case.

  For the N-model, we evaluate policies and compute the optimal benchmark using finite queue-length limits. Increasing these limits from 600 to 800 leaves the costs and optimality gaps essentially unchanged. We report optimality gaps only for policies that pass this convergence check.

The reentrant-line instances are too large for exact evaluation, so we estimate their long-run average holding costs by simulation. For the main results, we average 20 independent runs, each with 2 million events after a 200,000-event warm-up. Policies are evaluated using the same random-number streams.

For the six reentrant-line networks, the primary comparator is the PPO policy of
\citet{dai2022queueing}, evaluated on the same Markov chain we simulate. 
We also re-simulate LBFS and $c\mu$ as secondary checks.

\subsection{Protocol cross-check}
\label{app:q-qgym}

As a robustness check, we also evaluate the reentrant-line level-1 artifact under QGym's
finite-horizon protocol \citep{chen2024qgym}, which starts empty and averages $100$ trajectories.
\Cref{tab:q-qgym-crosscheck} shows that the level-1 comparison agrees with the steady-state
protocol.

\begin{table}[H]
\centering
\small
\input{tables/results_queueing_qgym_crosscheck}
\caption{Level-1 queueing protocol cross-check on the six reentrant-line instances. Entries are
average holding costs under QGym's finite-horizon protocol, which starts empty and averages $100$
trajectories of the first $T$ events. The gap is relative to the best of QGym's reported PPO-WC and
A2C-WC values; negative values are better and appear in \textbf{bold}.}
\label{tab:q-qgym-crosscheck}
\end{table}

\subsection{Timing}
\label{app:q-timing}

Queueing level-1 generation times vary with instance size: \model{gpt-5.6-sol} uses $45$--$1558$
seconds of sandbox compute on criss-cross, $2024$ seconds on the N-model, and usually most of the
$3600$-second budget on the reentrant-line instances. For L2, the criss-cross, N-model, and reentrant-line artifacts use $841$, $58$, and $208$
seconds of sandbox compute,
respectively. Returned \texttt{policy} calls are well under one millisecond. The criss-cross and
N-model \texttt{design} calls can slightly exceed the stated $30$-second target because they solve
small dynamic programs.

\input{app_queueing_mechanism}

\section{Additional assortment results}
\label{app:full-assortment}

This appendix gives additional details for the assortment results in
\Cref{sec:assortment}, including entry selection, the level-1 counterpart, and compute times.

\paragraph{Instance and entry selection.}
No instance is excluded from the MMNL or nested-logit comparisons: every released instance
($628$ MMNL, $971$ nested logit) is scored for every method. For constrained MMNL, we use the
authors' $1{,}800$-instance grid after dropping six degenerate instances with no feasible product.
A small number of individual \emph{entries} in the authors' per-instance results are excluded:
on MMNL, $27$ conic-solver revenues exceed the verified exact optimum. We exclude any author entry that exceeds the verified
optimum by more than a relative tolerance of $10^{-5}$.
LLM revenues are computed by our own evaluator on the
same instances, are checked for feasibility directly, and are never excluded or capped.

For level-1 assortment, one query is needed for each evaluated instance, so we use a
smaller stratified set. We take the lowest-seed instance from every released MMNL and nested-logit
configuration, giving $72$ MMNL and $48$ nested-logit instances, and the two lowest seeds from each
of the $18$ constrained-MMNL configurations, giving $36$ constrained-MMNL instances.

\begin{figure}[H]
\centering
\includegraphics[width=\linewidth]{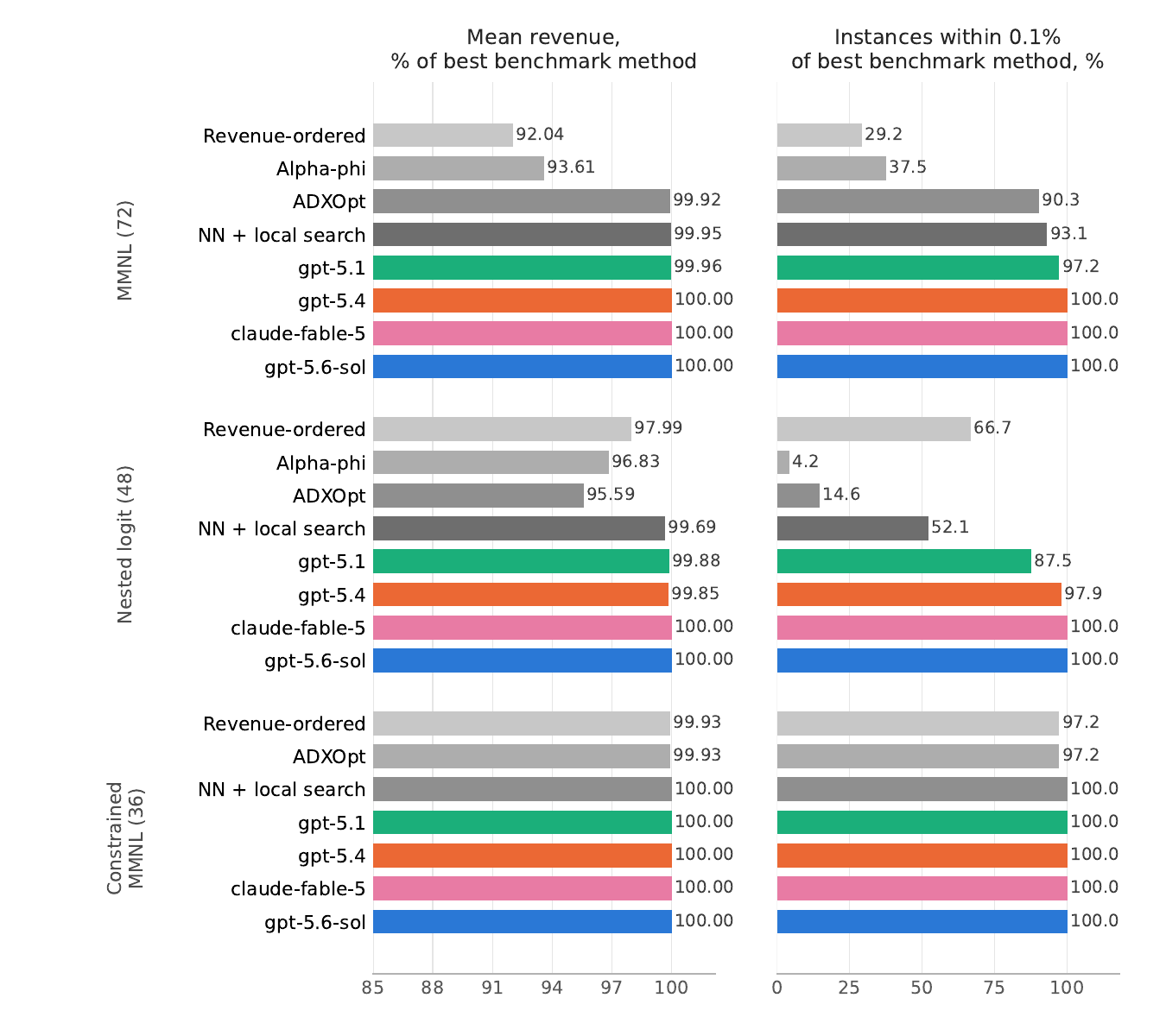}
\caption{Level-1 assortment results on the stratified instance set. This is the level-1 counterpart
of \Cref{fig:assortment-vs-published}: one query per instance, with existing methods scored on the
same instances. The comparator, axes, and conventions match \Cref{fig:assortment-vs-published}. 
}
\label{fig:assortment-vs-published-l1}
\end{figure}

\subsection{Compute and wall time}
\label{app:assortment-timing}

\Cref{tab:assortment-timing-l2,tab:assortment-timing-l1} report three clocks for
\model{gpt-5.6-sol}: query wall time, sandbox Python time, and, for level 2, the runtime of the
returned \texttt{solve} function on one evaluation instance.

Level-2 queries use about $100$ to $120$ s of their $3{,}600$ s sandbox budget, and the returned
solvers usually run in well under a second per instance. Zero-compute queries take about four
minutes of reasoning and return solvers of similar speed. Level-1 queries are usually about a
minute, with a few MMNL and nested-logit queries using most of the $900$ s sandbox budget.

\begin{table}[H]
\centering
\small
\input{tables/results_assortment_timing_l2}
\caption{Level-2 assortment queries and returned solvers, \model{gpt-5.6-sol}: the algorithm
used in the main results for each family (the first valid replicate) at the full budget, and the
zero-compute query of \Cref{sec:zero-compute}. Wall is the time from sending the prompt to the
final answer; Sandbox is the Python execution time used of the budget, over the number of code
cells run; output tokens include reasoning tokens. Solver runtime is per evaluation instance (mean
and maximum over the family), against the $30$ s target stated in the prompt; constrained MMNL is
measured on the authors' $1{,}794$ instances.}
\label{tab:assortment-timing-l2}
\end{table}

\begin{table}[H]
\centering
\small
\input{tables/results_assortment_timing_l1}
\caption{Level-1 assortment queries, \model{gpt-5.6-sol}: distribution over the level-1 instance
set (one query per instance) of query wall time, sandbox seconds used of the $900$ s budget, code
cells run, and output tokens including reasoning tokens.}
\label{tab:assortment-timing-l1}
\end{table}

\section{Holdout generalization results}
\label{app:l2-holdout-results}

This appendix gives additional details for the holdout check in \Cref{sec:l2-holdout}. The
level-2 artifacts are frozen from the main runs and evaluated on new instances; no model is queried
again. \Cref{fig:l2-holdout-inventory} reports the inventory results in the main text. For
assortment, \Cref{tab:l2-holdout-assortment} reports mean revenue ratios, with the minimum and
maximum across instances in brackets, versus the exact optimum where available and otherwise versus
the stated heuristic comparator.

\paragraph{Instance construction.}

\emph{Lost sales inventory.} The holdout has 30 deterministic instances and 12 stochastic
instances. The deterministic grid uses short lead times
$\ell\in\{2,3,4\}$ with $(\lambda,p)\in\{(7,6),(18,29)\}$, and longer lead times
$\ell\in\{6,9,12\}$ with $\lambda\in\{7,18,35,50\}$ and $p\in\{6,29\}$. The stochastic grid
uses $(\ell_{\min},\ell_{\max})\in\{(2,8),(3,10),(4,12)\}$,
$\lambda\in\{7,18\}$, and $p\in\{6,29\}$; each order's lead time is drawn uniformly from
$\{\ell_{\min},\ldots,\ell_{\max}\}$. All lost-sales holdouts use holding cost $h=1$, ordering
cost $c=0$, Poisson demand, and maximum order quantity $\lceil 2.5\lambda\rceil+2$.

\emph{MMNL.} The MMNL holdout replaces the released benchmark's two-utility structure
with continuous utilities. We use two regimes. In the iid-lognormal regime, prices are
$\exp(N(1.2,0.55^2))$, utilities are $\exp(N(0,0.9^2))$, and outside utilities are
$\exp(N(0.2,0.25^2))$. In the correlated-quality regime, product qualities and features are
standard normal, class tastes are $N(0,0.8^2)$, prices increase with quality, and utilities depend
on quality and feature-taste interactions. Class weights are Dirichlet with unit parameters. The
exact subset uses $n\in\{30,35\}$, $m\in\{5,10,25\}$, cap $k\in\{4,6\}$, both regimes, and two
seeds per configuration. The large subset uses $n\in\{100,200\}$, $m\in\{10,25\}$,
$k\in\{10,20\}$, and both regimes.

\emph{Nested logit.} The nested-logit holdout uses $n\in\{30,50\}$ products per nest,
$m\in\{10,20\}$ nests, and per-nest cap $k\in\{5,10,20\}$. Each instance draws
$\gamma_i\sim U[0.35,0.85]$, $v_{i0}\sim\exp(N(0,0.35^2))$, and
$v_0\sim\exp(N(1.2,0.35^2))$. We again use two regimes. In the anti-correlated regime, high-utility
products tend to have lower prices. In the cross-nest regime, nests differ in overall attraction
and price level, and products then vary within each nest.

\begin{table}[!htbp]
\centering
\input{tables/results_l2_holdout_assortment}
\caption{Assortment holdout results for frozen level-2 algorithms. Entries are mean revenue ratios, with the minimum and maximum over holdout instances in brackets; larger is better.}
\label{tab:l2-holdout-assortment}
\end{table}

\section{Adding a web-search tool}
\label{app:websearch}

The main experiments give the model no retrieval tool: its only tool is the Python sandbox, and
none of the executed code accessed the network. As a variant, we reran four \model{gpt-5.6-sol}
level-2 queries (deterministic lost sales, MMNL, nested logit, and constrained MMNL) with
OpenAI's built-in web-search tool attached, under the same $3600$-second compute budget and output
contract. The prompt is unchanged except for one appended paragraph:

\begin{quote}\small\itshape
You also have a web search tool. You may use it as much as you like; there is no limit on the
number of searches. Web searches do not count against the compute budget, which applies to Python
execution only.
\end{quote}

Each run produces one algorithm, evaluated as in the main experiments. Every search query and page
visit is logged. \Cref{tab:websearch-activity} shows that the model used the tool sparingly, and only at the start of a session.

\begin{table}[H]
\centering
\small
\begin{tabular}{lp{0.3\textwidth}p{0.42\textwidth}}
\toprule
Problem class & Search activity & Pages opened \\
\midrule
Lost sales, deterministic & none (tool never invoked) & --- \\
MMNL & one query, first turn:\newline \emph{``mixed multinomial logit assortment optimization
approximation algorithm paper''} & the mixture-of-logits paper of
\citet{rusmevichientong2014random}; the conic-optimization paper of \citet{sen2018conic}
behind the exact MMNL solver; a text search for ``Algorithm 1'' in a preprint \\
Nested logit & one query, first turn:\newline \emph{``nested logit assortment optimization
polynomial algorithm revenue ordered assortments per nest''} & the constrained nested-logit
paper of \citet{gallego2014constrained} (twice) \\
Constrained MMNL & none (tool never invoked) & --- \\
\bottomrule
\end{tabular}
\caption{Web-search activity of the four \model{gpt-5.6-sol} level-2 queries that were given
the search tool. All searches occurred in the first turn, before any code was written.}
\label{tab:websearch-activity}
\end{table}

The results are essentially unchanged. Deterministic lost sales has mean gap $-1.26\%$ versus the
best benchmark method, compared with $-1.27\%$ for the main artifact. MMNL again attains the exact
optimum on all $628$ instances, constrained MMNL again matches the best benchmark method on all
$1{,}794$ instances, and nested logit matches the main artifact.

Retrieval does not appear to be the binding constraint in these runs. When the model searches, it
goes to canonical references, but the returned algorithms and results do not change. This is a
limited check, using one artifact per class and one model; the web-search tool also increased
wall-clock generation time substantially.

\clearpage

\section{How the model reasons}
\label{app:reasoning}

This appendix summarizes selected full-budget \model{gpt-5.6-sol} level-2 transcripts from
the inventory, assortment, and queueing experiments. The logs include provider-generated \emph{reasoning summaries}: short
first-person summaries of what the model was doing before it wrote the final algorithms.

Across these transcripts, the same pattern appears: the model identifies structure, compares
algorithm families, budgets computation, builds small
testbeds, adds guardrails, and formats the returned code. We illustrate with representative
quotes; turn numbers refer to the transcript of the named problem class.

\paragraph{1. Formalize the problem and recall what is known.}
The first turn usually brings in a relevant structural result or known algorithmic
idea.
\begin{itemize}[leftmargin=*, itemsep=2pt, topsep=3pt]
\item Nested logit, turn~1: \emph{``I'm considering the revenue-order theorem \ldots\ It seems the optimal
assortment is revenue-ordered globally for Multinomial Logit (MNL) models, but not for nested logit
which operates differently. For nested logit, I'm thinking it might be more about enumerating the
top $k$ in each nest.''}
\item Lost sales, turn~2: \emph{``I really want to compare optimized base stock versus the current
methods, as literature suggests that the latter performs better in terms of lost sales. A hybrid
approach might be beneficial too!''}
\end{itemize}

\paragraph{2. Enumerate candidate method families and weigh exactness against cost.}
Before writing code, the model compares exact methods, simple policies, and hybrids.
\begin{itemize}[leftmargin=*, itemsep=2pt, topsep=3pt]
\item Lost sales, turn~2: \emph{``I'm considering calculating the exact distribution dynamics, but using
distributions support and scipy convolution might be too heavy for each call. Instead, I could
approximate projections with fluid dynamics while calibrating safety corrections based on state
features. Another option might be employing an inventory position policy, which is simpler and
potentially exact.''}
\item Multi-echelon, turn~1: \emph{``The cost surface looks mostly convex, so using analytic seeds with a
coarse-to-fine simulation should help refine the results \ldots\ I wonder if a simpler proportional
allocation could work better!''}
\end{itemize}

\paragraph{3. Budget the computation.}
The model does back-of-the-envelope runtime arithmetic against the stated $30$-second
per-instance target and the capabilities of the sandbox.
\begin{itemize}[leftmargin=*, itemsep=2pt, topsep=3pt]
\item Dual sourcing, turns~1--2: \emph{``I'm considering the cost of simulating 500 candidates across 100k
steps \ldots\ each evaluation involves around 800 tiny numpy calls, likely taking about 0.007
seconds''}; \emph{``I need to consider vectorization or JIT compilation, but that's unavailable.''}
\item Constrained MMNL, turn~6: \emph{``I'm considering time scheduling for around 30 seconds, using
12--18 seconds for MILP and the rest for heuristics.''}
\item Criss-cross queueing, turn~8: \emph{``I'm digging into sparse policy iteration for a setup
with around 200k states \ldots\ I also need to address potential issues at finite boundaries and
examine policy action extraction.''}
\end{itemize}

\paragraph{4. Build a testbed and benchmark against ground truth.}
The model writes simulators, brute-force checkers, dynamic programs, or internal testbeds inside
the sandbox. It uses exact solutions on small cases as a yardstick, constructs hard test cases, and
watches for simulation noise. In dual sourcing, it keeps the exact dynamic program in the returned
algorithm when the state space is small enough. The returned algorithm has a mean cost gap of $0.03\%$ across the six instances.
\begin{itemize}[leftmargin=*, itemsep=2pt, topsep=3pt]
\item Stochastic lead times, turn~5: \emph{``I think implementing an exact case with $b=2$ could yield a
1--2\% improvement \ldots\ build an exact Value Iteration (VI) model to compare selected policies and
assess any gaps.''}
\item MMNL, turn~2: \emph{``I'm looking at a random landscape where easy typical structures exist, but I
need adversarial mixtures for trapping locals.''}
\item Reentrant-line queueing, turn~3: \emph{``If I'm considering, say, 300,000 iterations,
there's a concern about Python loop overhead taking around 2 seconds.''}
\end{itemize}

\paragraph{5. Hunt for failure modes and add guardrails.}
Much of the later reasoning is about class-level edge cases. These checks become explicit
branches in the returned code: never-order and maximum-order cases, overflow guards, and
feasibility repair.
\begin{itemize}[leftmargin=*, itemsep=2pt, topsep=3pt]
\item Multi-echelon, turn~4: \emph{``setting the stock at zero means the manufacturer won't place orders,
even if stores need inventory. If the warehouse is empty, anticipated shipments become zero, leading
to a perpetual lack of stock. I need to establish an ordering policy that anticipates future store
demand.''}
\item Lost sales, turn~7, on its own tuned parameter range: \emph{``The gamma bounds are between $-3$ and
$2$, but that might lead to strange out-of-distribution states.''}
\end{itemize}

\paragraph{6. Finalize to the contract.}
The last turn is mostly about the required interface: the function signature, imports,
and behavior on degenerate inputs.
\begin{itemize}[leftmargin=*, itemsep=2pt, topsep=3pt]
\item MMNL, turn~9: \emph{``The user wants the final output to be just the source code.''}
\item Nested logit, turn~10: the model checks that its scaling of the nest utilities \emph{``never raises
issues on valid shapes or ranges.''}
\end{itemize}

\section{Prompts}
\label{app:prompts}

The experimental prompts are generated from shared prompt code and frozen in the repository. This
appendix reports representative examples: one level-1 and one level-2 prompt for inventory and
assortment, and one level-1 prompt plus L2 and Broad L2 prompts for queueing. For exact
replication, use the full frozen prompts in the public repository.\footnote{\url{https://anonymous.4open.science/r/llm-or-algorithms-F9F2}.}
The same repository also holds the complete prompt files and run records: every LLM query with its reasoning summaries, executed code, outputs, and token usage.

\lstdefinestyle{promptstyle}{
  language={},
  basicstyle=\ttfamily\scriptsize,
  breaklines=true,
  showstringspaces=false,
  numbers=none,
  xleftmargin=0pt,
  frame=single
}
\providecommand{\promptheading}[1]{\par\addvspace{0.8em}\noindent\textbf{#1}\par\nopagebreak\vspace{0.25em}}

\input{prompts_appendix_examples}

%% file: tables/results_dual_sourcing_llm.tex
\small\setlength{\tabcolsep}{4.5pt}
\begin{tabular}{lrrrrrrrrr}
\toprule
 & Optimum & \multicolumn{2}{c}{\model{gpt-5.6-sol}} & \multicolumn{2}{c}{\model{claude-fable-5}} & \multicolumn{2}{c}{\model{gpt-5.4}} & \multicolumn{2}{c}{\model{gpt-5.1}} \\
Instance & & L1 & L2 & L1 & L2 & L1 & L2 & L1 & L2 \\
\midrule
$\ell_r=2,\,c_e=105$ & 216.75 & \textbf{+0.00} & \textbf{+0.00} & \textbf{+0.00} & \textbf{+0.00} & \textbf{+0.00} & \textbf{+0.06} & +0.17 & +0.55 \\
$\ell_r=2,\,c_e=110$ & 219.71 & \textbf{+0.00} & \textbf{+0.00} & \textbf{+0.00} & \textbf{+0.00} & \textbf{+0.00} & +0.17 & +1.51 & +0.69 \\
$\ell_r=3,\,c_e=105$ & 216.85 & \textbf{+0.00} & \textbf{+0.00} & \textbf{+0.00} & \textbf{+0.00} & \textbf{+0.00} & +0.16 & +0.63 & +0.63 \\
$\ell_r=3,\,c_e=110$ & 220.32 & \textbf{+0.00} & \textbf{+0.00} & \textbf{+0.00} & \textbf{+0.00} & \textbf{+0.00} & +0.11 & +0.76 & +0.83 \\
$\ell_r=4,\,c_e=105$ & 216.87 & \textbf{+0.00} & \textbf{+0.00} & \textbf{+0.00} & \textbf{+0.00} & \textbf{+0.00} & +0.23 & \textbf{+0.00} & +0.71 \\
$\ell_r=4,\,c_e=110$ & 220.58 & \textbf{+0.00} & +0.16 & \textbf{+0.00} & \textbf{+0.00} & \textbf{+0.00} & +0.13 & +0.92 & +0.96 \\
\bottomrule\end{tabular}

%% file: tables/results_multi_echelon.tex
\small\setlength{\tabcolsep}{3.8pt}
\begin{tabular}{lrrrrrrrrr}
\toprule
 & Benchmark & \multicolumn{2}{c}{\model{gpt-5.6-sol}} & \multicolumn{2}{c}{\model{claude-fable-5}} & \multicolumn{2}{c}{\model{gpt-5.4}} & \multicolumn{2}{c}{\model{gpt-5.1}} \\
Instance & & L1 & L2 & L1 & L2 & L1 & L2 & L1 & L2 \\
\midrule
\makecell[l]{Setting 1\\ \footnotesize$(\ell_w{=}2,\,\ell_r{=}2,\,\mu{=}5,\,\sigma{=}14)$} & 906.48 & \textbf{-22.06} & \textbf{-12.18} & \textbf{-20.14} & \textbf{-14.91} & \textbf{-9.13} & \textbf{-19.94} & +1.32 & +3.84 \\
\makecell[l]{Setting 2\\ \footnotesize$(\ell_w{=}5,\,\ell_r{=}3,\,\mu{=}0,\,\sigma{=}20)$} & 1138.42 & \textbf{-23.50} & \textbf{-10.25} & \textbf{-18.63} & \textbf{-17.19} & \textbf{-4.65} & \textbf{-21.43} & \textbf{-0.34} & +0.20 \\
\bottomrule\end{tabular}

%% file: tables/results_lost_sales.tex
\footnotesize\setlength{\tabcolsep}{2.6pt}
\begin{tabular}{lrrrrrrrrrrrr}
\toprule
 & \multicolumn{4}{c}{Benchmarks} & \multicolumn{2}{c}{\model{gpt-5.6-sol}} & \multicolumn{2}{c}{\model{claude-fable-5}} & \multicolumn{2}{c}{\model{gpt-5.4}} & \multicolumn{2}{c}{\model{gpt-5.1}} \\
Instance & CO & BS & CBS & Mixed & L1 & L2 & L1 & L2 & L1 & L2 & L1 & L2 \\
\midrule
\multicolumn{13}{l}{\emph{Lead-time sweep ($\lambda=5$, $p=4$)}} \\
$\ell=5$ & 5.272 & 5.371 & \underline{4.935} & 5.128 & \textbf{-2.50} & \textbf{-2.45} & \textbf{-2.50} & \textbf{-2.50} & +8.83 & +0.00 & +8.83 & +19.51 \\
$\ell=6$ & 5.272 & 5.510 & \underline{5.033} & 5.128 & \textbf{-3.23} & \textbf{-3.08} & \textbf{-3.23} & \textbf{-3.23} & \textbf{-3.22} & +0.38 & +18.47 & +18.47 \\
$\ell=7$ & 5.272 & 5.629 & \underline{5.123} & 5.129 & \textbf{-4.06} & \textbf{-4.06} & \textbf{-4.10} & +0.00 & \textbf{-3.86} & +0.95 & +16.90 & +17.51 \\
$\ell=8$ & 5.272 & 5.722 & 5.189 & \underline{5.129} & \textbf{-3.46} & \textbf{-3.56} & \textbf{-3.59} & +1.18 & \textbf{-1.96} & +3.13 & +11.56 & +18.13 \\
$\ell=9$ & 5.272 & 5.801 & 5.253 & \underline{5.129} & \textbf{-3.11} & \textbf{-3.08} & \textbf{-3.09} & +2.42 & +10.84 & +2.82 & +13.11 & +18.93 \\
$\ell=10$ & 5.272 & 5.870 & 5.271 & \underline{5.130} & \textbf{-2.78} & \textbf{-2.73} & \textbf{-2.78} & +2.76 & +2.76 & +2.78 & \textbf{-1.90} & +19.57 \\
\midrule
\multicolumn{13}{l}{\emph{Demand scale ($\ell=8$, $p=4$)}} \\
$\lambda=10$ & 7.242 & 8.174 & \underline{7.084} & 7.237 & \textbf{-0.46} & \textbf{-0.41} & \textbf{-0.33} & -0.03 & \textbf{-0.43} & +0.12 & +15.40 & +29.70 \\
$\lambda=15$ & 9.322 & 10.061 & \underline{8.687} & 8.950 & \textbf{-0.32} & -0.06 & +0.01 & -0.07 & +0.19 & +0.00 & +15.82 & +37.69 \\
$\lambda=20$ & 10.739 & 11.649 & \underline{10.085} & 10.348 & \textbf{-0.60} & \textbf{-0.39} & \textbf{-0.55} & +0.00 & +0.00 & +0.00 & +15.50 & +42.16 \\
$\lambda=25$ & 11.688 & 13.031 & \underline{11.326} & 11.563 & \textbf{-0.98} & \textbf{-0.97} & +23.31 & +0.00 & \textbf{-0.69} & +0.00 & +15.06 & +49.20 \\
\midrule
\multicolumn{13}{l}{\emph{Penalty cost ($\ell=8$, $\lambda=5$)}} \\
$p=9$ & 10.277 & 8.325 & \underline{7.542} & 8.348 & \textbf{-0.90} & \textbf{-0.76} & \textbf{-0.60} & +0.00 & +0.38 & +2.44 & +20.65 & +20.65 \\
$p=19$ & 20.286 & 10.916 & \underline{10.378} & 12.745 & \textbf{-1.95} & \textbf{-2.01} & \textbf{-1.95} & +0.00 & +3.68 & +3.07 & +5.18 & +19.50 \\
$p=49$ & 50.312 & 14.169 & \underline{13.744} & 21.149 & \textbf{-0.68} & \textbf{-0.79} & \textbf{-0.93} & +0.00 & +2.05 & +2.64 & +3.09 & +26.10 \\
\midrule
\multicolumn{13}{l}{\emph{Stochastic lead times ($\lambda=5$, $p=4$, $\ell_{\min}=2$)}} \\
$\bar\ell=5$ & 7.191 & 7.294 & \underline{7.027} & -- & \textbf{-2.04} & \textbf{-0.42} & \textbf{-2.05} & \textbf{-1.47} & \textbf{-1.51} & +3.79 & +3.79 & +3.79 \\
$\bar\ell=6$ & 7.484 & 7.640 & \underline{7.355} & -- & \textbf{-1.22} & \textbf{-0.64} & \textbf{-2.21} & \textbf{-1.70} & \textbf{-1.82} & +3.88 & +3.35 & +3.88 \\
$\bar\ell=7$ & 7.725 & 7.915 & \underline{7.603} & -- & \textbf{-2.14} & \textbf{-0.74} & \textbf{-2.26} & \textbf{-1.80} & +0.45 & +4.14 & +3.73 & +4.14 \\
$\bar\ell=8$ & 7.923 & 8.138 & \underline{7.789} & -- & \textbf{-2.07} & \textbf{-0.71} & \textbf{-2.10} & \textbf{-1.74} & +3.18 & +4.48 & +4.48 & +4.61 \\
$\bar\ell=9$ & 8.110 & 8.332 & \underline{7.945} & -- & \textbf{-1.92} & \textbf{-0.64} & \textbf{-1.27} & \textbf{-1.64} & +4.70 & +4.87 & +4.37 & +4.87 \\
$\bar\ell=10$ & 8.267 & 8.500 & \underline{8.069} & -- & \textbf{-1.75} & \textbf{-0.48} & \textbf{-1.57} & \textbf{-1.42} & \textbf{-1.45} & +5.69 & +5.34 & +5.69 \\
$\bar\ell=11$ & 8.414 & 8.657 & \underline{8.187} & -- & +0.30 & \textbf{-0.55} & \textbf{-1.31} & \textbf{-1.38} & +9.17 & +5.74 & +5.37 & +5.74 \\
\bottomrule\end{tabular}

%% file: tables/results_inventory_timing_l2.tex
\begin{tabular}{@{}llrrrrrr@{}}
\toprule
 & & \multicolumn{4}{c}{Query} & \multicolumn{2}{c}{\texttt{design} runtime (s)} \\
\cmidrule(lr){3-6}\cmidrule(lr){7-8}
Class & Budget & Wall (s) & Sandbox (s) & Cells & Output tokens & Mean & Max \\
\midrule
Lost sales, deterministic & 3{,}600 s & 598 & 54.9 & 6 & 21,921 & 3.0 & 4.8 \\
Lost sales, stochastic & 3{,}600 s & 795 & 125.2 & 6 & 25,828 & 12.3 & 12.9 \\
Dual sourcing & 3{,}600 s & 828 & 237.0 & 13 & 25,223 & 4.8 & 11.9 \\
Multi-echelon & 3{,}600 s & 575 & 17.7 & 4 & 22,935 & 11.1 & 13.1 \\
\midrule
Lost sales, deterministic & 0 s & 320 & 0.0 & 0 & 14,167 & 4.0 & 7.3 \\
\bottomrule
\end{tabular}

%% file: tables/results_inventory_timing_l1.tex
\begin{tabular}{@{}lrrrrrrrrr@{}}
\toprule
 & \multicolumn{3}{c}{Wall (s)} & \multicolumn{3}{c}{Sandbox (s)} & Output & \multicolumn{2}{c}{Module load (s)} \\
\cmidrule(lr){2-4}\cmidrule(lr){5-7}\cmidrule(lr){9-10}
Class (\# queries) & Mean & SD & Max & Mean & SD & Max & tokens & Mean & Max \\
\midrule
Lost sales, deterministic (19) & 2,248 & 1,508 & 4,040 & 1,959 & 1,432 & 3,600 & 14,559 & 1.8 & 25.7 \\
Lost sales, stochastic (7) & 3,205 & 360 & 3,492 & 2,841 & 393 & 3,190 & 19,438 & 0.2 & 0.5 \\
Dual sourcing (6) & 240 & 87 & 370 & 37 & 45 & 129 & 10,330 & 1.2 & 3.5 \\
Multi-echelon (2) & 3,134 & 677 & 3,811 & 2,661 & 818 & 3,479 & 27,148 & 0.0 & 0.0 \\
\midrule
Lost sales, lead-time sweep, 0 s (6) & 203 & 25 & 238 & 0 & 0 & 0 & 9,364 & 0.4 & 1.9 \\
\bottomrule
\end{tabular}

%% file: tables/results_zero_compute_usage.tex
\begin{tabular}{lllrrrrrr}
\toprule
\textbf{Level} & \textbf{Compute} & \textbf{Measure} & $\ell=5$ & $\ell=6$ & $\ell=7$ & $\ell=8$ & $\ell=9$ & $\ell=10$ \\
\midrule
\multirow{5}{*}{Level 1} & \multirow{3}{*}{3600 s} & sandbox compute & 801 & 3,100 & 3,563 & 3,504 & 2,400 & 3,600 \\
 & & wall clock & 1,137 & 3,404 & 3,820 & 3,772 & 2,711 & 4,040 \\
 & & module load & 6.9 & 0.5 & 0.2 & 0.1 & 0.0 & 25.7 \\
\cmidrule(l){2-9}
 & \multirow{2}{*}{0 s} & wall clock & 222 & 212 & 168 & 238 & 173 & 202 \\
 & & module load & 0.1 & 1.9 & 0.1 & 0.1 & 0.1 & 0.1 \\
\midrule
\multirow{5}{*}{Level 2} & \multirow{3}{*}{3600 s} & sandbox compute & \multicolumn{6}{c}{55 (one query, whole class)} \\
 & & wall clock & \multicolumn{6}{c}{598 (one query, whole class)} \\
 & & \texttt{design} & 2.5 & 2.9 & 3.2 & 3.8 & 4.5 & 4.8 \\
\cmidrule(l){2-9}
 & \multirow{2}{*}{0 s} & wall clock & \multicolumn{6}{c}{320 (one query, whole class)} \\
 & & \texttt{design} & 3.2 & 3.8 & 4.0 & 7.3 & 4.7 & 4.6 \\
\bottomrule\end{tabular}

%% file: app_inventory_mechanism.tex
\subsection{Projected-inventory mechanism experiment}
\label{app:inventory-mechanism}

This experiment refits PIL and the main deterministic level-2 rule from \model{gpt-5.6-sol} on
the $19$ settings of \Cref{tab:inv-instances}, to measure how much the standard-deviation term
adds. These fits are separate from the original LLM runs and do not change any main-text result.

\paragraph{Projection.}
Both rules order based on the stock $J$ left just before the new order arrives. Starting from
on-hand stock $I$ and scheduled receipts $a_1,\ldots,a_{\ell-1}$, with independent
$\mathrm{Poisson}(\lambda)$ demands $D_j$,
\[
R_0=(I-D_0)^+,\qquad R_j=(R_{j-1}+a_j-D_j)^+,\qquad J=R_{\ell-1}.
\]
We compute the mean and variance of $J$ in two ways. The \emph{normal engine} is the LLM's own
method: it computes the first step exactly and then approximates each later step with a normal
distribution floored at zero. The \emph{accurate engine} propagates the full discrete distribution
of $J$, with no normal approximation. Both rules round the order to the nearest integer and clip it
to $[0,\bar q]$, where $\bar q=\lceil2.5\lambda\rceil+2$ is fixed by the instance.

\paragraph{Calibration.}
For each setting, we fit PIL's target $T$ (with $\gamma=0$) and the extended rule's $(T,\gamma)$
separately, using the same simulation search: a coarse grid over $T$ and $\gamma\in[-3,2]$, then
local refinement around the best candidates with longer simulations and fresh demand paths at each
stage. The extended rule's search includes the fitted PIL policy as a candidate, so it can always
fall back to PIL.

\paragraph{Evaluation.}
All fitted policies are evaluated on the same $60$ fresh demand paths per setting, each with
$30{,}000$ measured periods after a $3{,}000$-period warmup. For each setting we report the
percentage cost change $100(C_A/C_B-1)$, and the grid effect averages this over the $19$ settings.
Confidence intervals are paired across paths, and per-setting significance uses a Holm correction
across the $19$ settings. Each policy is fitted once, so the intervals do not include fitting
variability. The primary standard-deviation contrasts use PIL as the baseline, so they are not comparable to the main-text
gaps, which use the best benchmark method.

\begin{table}[H]
\centering
\small
\input{tables/results_inventory_mechanism_summary}
\caption{Mechanism contrasts: mean relative cost change, with paired 95\%
confidence intervals conditional on the fitted policies. Each row uses its
specified comparison policy as denominator. Negative values favor the first
policy; the first two rows use the PIL fit from the same projection engine.}
\label{tab:inventory-mechanism-summary}
\end{table}

\paragraph{Results.}
\Cref{tab:inventory-mechanism-summary} reports the average effects, and
\Cref{fig:inventory-pil-mechanism} shows the change on each setting. With either engine, the extended rule beats its
own PIL fit on every setting after Holm correction, and every fitted $\gamma$ is negative. The two
engines give nearly identical costs, so the normal approximation has little effect. Refitting the
LLM's rule lowers its cost by only $0.048\%$ on average relative to its original parameters, so the
original \texttt{design} routine already tuned it well.

We checked our reimplementation of the LLM rule against the original saved policy, and the
compiled simulator against our Python simulator, on all $19$ settings. The code, seeds, and full
results are in the code repository.

%% file: tables/results_inventory_mechanism_summary.tex
\begin{tabular}{lrr}
\toprule
Comparison & Mean change & 95\% interval \\
\midrule
Normal two-parameter vs. normal PIL & $-0.447\%$ & $[-0.456,\,-0.438]\%$ \\
Accurate two-parameter vs. accurate PIL & $-0.445\%$ & $[-0.454,\,-0.436]\%$ \\
Accurate PIL vs. normal PIL & $-0.004\%$ & $[-0.009,\,+0.001]\%$ \\
Accurate two-parameter vs. normal two-parameter & $-0.003\%$ & $[-0.008,\,+0.003]\%$ \\
Retuned normal two-parameter vs. original & $-0.048\%$ & $[-0.055,\,-0.041]\%$ \\
\bottomrule
\end{tabular}

%% file: tables/results_queueing_qgym_crosscheck.tex
\begin{tabular}{lrrrr}
\toprule
\textbf{Instance} & \textbf{$T$ events} & \textbf{L1 cost} & \textbf{Best QGym RL} & \textbf{Gap} \\
\midrule
$L=2$ stations, $6$ queues & 10{,}000 & $12.90 \pm 0.44$ & $13.0$ & $\mathbf{-0.9\%}$ \\
$L=3$ stations, $9$ queues & 50{,}000 & $20.54 \pm 0.33$ & $22.0$ & $\mathbf{-6.6\%}$ \\
$L=4$ stations, $12$ queues & 80{,}000 & $29.18 \pm 0.36$ & $29.7$ & $\mathbf{-1.8\%}$ \\
$L=5$ stations, $15$ queues & 100{,}000 & $39.03 \pm 0.55$ & $38.7$ & $+0.8\%$ \\
$L=6$ stations, $18$ queues & 200{,}000 & $41.74 \pm 0.37$ & $47.4$ & $\mathbf{-11.9\%}$ \\
$L=7$ stations, $21$ queues & 100{,}000 & $55.68 \pm 0.92$ & $56.3$ & $\mathbf{-1.1\%}$ \\
\bottomrule
\end{tabular}

%% file: app_queueing_mechanism.tex
\subsection{Reentrant-line pressure mechanism experiment}
\label{app:queueing-mechanism}

This experiment compares the LLM's reentrant-line score with the Dai--Lin family on the six
reentrant lines ($L=2,\ldots,7$), with unit holding costs and nominal station utilization $0.9$.
These fits are separate from the original LLM runs and do not change the main-text results.

\paragraph{Relationship to Dai--Lin.}
With buffer levels $P_q=w_qX_q-\theta_q$, the Dai--Lin pressure expands as
\[
\mu_q(P_q-P_{r(q)})
=\mu_q(w_qX_q-w_{r(q)}X_{r(q)})+\mu_q(\theta_{r(q)}-\theta_q).
\]
Setting $\theta_q=\theta_{r(q)}-B_q/\mu_q$ backward from the exit, where $\theta=0$, produces any
bonuses $B_q$, so the LLM score with $\gamma=1$ is a Dai--Lin score. A discount $\gamma<1$
generally cannot be absorbed into the weights.

\paragraph{Five policies.}
All policies use the score \eqref{eq:queueing-llm-pressure}.
\begin{enumerate}[leftmargin=*,itemsep=2pt,topsep=4pt]
\item \textbf{Basic maximum pressure:} $w_q=1$, $B_q=0$, $\gamma=1$; idling permitted.
\item \textbf{Weighted, non-idling:} tuned $w_q$ and $B_q$, $\gamma=1$.
\item \textbf{Weighted, idling permitted (Dai--Lin):} tuned $w_q$ and $B_q$, $\gamma=1$.
\item \textbf{Free discount, non-idling (LLM form):} tuned $w_q$, $B_q$ and $\gamma\in[0,1]$.
\item \textbf{Free discount, idling permitted:} tuned $w_q$, $B_q$ and $\gamma\in[0,1]$.
\end{enumerate}
For Policies~2--5, we tune the parameters continuously over a much wider range than the LLM's
original finite menu, so Policy~4 tests the LLM's policy structure, not its original parameter
choices. Comparing Policies~1 and~3 isolates the value of tuned weights and bonuses, comparing~2
with~4 (or~3 with~5) tests discounting, and comparing~2 with~3 (or~4 with~5) tests idling.

\paragraph{Fitting.}
We fit each policy's parameters by simulation, using a derivative-free evolutionary search (CMA-ES; \citealp{hansen2016cma}), starting from the LLM's original menu
policies among other points, then refine on longer simulations and select on fresh paths. We keep
searching from the selected policy until a final check finds less than $0.25\%$ further gain on
independent paths. Candidates found for one family are shared with the families that contain it.
A non-idling policy can be reproduced on bounded queue lengths by adding a large common bonus at
each station, so the fitted non-idling rule is also a candidate for the idle-permitted families.
The search does not certify global optimality.

\paragraph{Evaluation.}
After selection, all five policies are evaluated on the same $40$ fresh paths per network, each with
eight million simulated events after warmup. The simulator is uniformized, so all policies see the
same random events, and it is checked against an independent Python implementation.
We report trajectory-level $t$ confidence intervals conditional on the fitted policies.

\paragraph{Results.}
\Cref{tab:queueing-mechanism-costs} reports the costs. Tuning
weights and bonuses (Policy~3) lowers mean cost by $42.2\%$--$76.5\%$ relative to basic maximum
pressure. Policies~2--5 differ by at most $0.25\%$ in mean cost on every network. For
$L=4,\ldots,7$, the free-discount fits choose $\gamma=1$, and all four policies give identical
costs on every path; at $L=2,3$ they choose $\gamma\approx0.87$ and $0.96$, with small
differences. Permitting idling also makes no material difference: Policies~2 and~3 give identical
costs on all $240$ paths; voluntary idling is observed only very rarely for Policy~5 at $L=3$.
So tuned Dai--Lin policies
essentially match the performance of the tuned LLM variant on these instances, and we find no
material additional benefit from discounting or from forced service.

\begin{table}[H]
\centering
\small
\setlength{\tabcolsep}{3pt}
\input{tables/results_queueing_mechanism_costs}
\caption{Fresh mean holding costs $\pm$ pointwise $95\%$ confidence-interval
half-widths, from $40$ independent paired paths per network. Policy numbers
refer to the definitions above. Weights and bonuses, and $\gamma$
when free, are continuously fitted with eligible incumbent sharing.
These follow-up results do not replace \Cref{tab:q-main-results}.}
\label{tab:queueing-mechanism-costs}
\end{table}

%% file: tables/results_queueing_mechanism_costs.tex
\begin{tabular}{rrrrrr}
\toprule
$L$ & \makecell{1. Basic\\pressure} & \makecell{2. $\gamma=1$\\non-idling} & \makecell{3. $\gamma=1$\\idle permitted} & \makecell{4. Free $\gamma$\\non-idling} & \makecell{5. Free $\gamma$\\idle permitted} \\
\midrule
2 & $23.333\pm0.186$ & $13.497\pm0.100$ & $13.497\pm0.100$ & $13.531\pm0.100$ & $13.531\pm0.100$ \\
3 & $49.113\pm0.335$ & $21.324\pm0.144$ & $21.324\pm0.144$ & $21.293\pm0.150$ & $21.293\pm0.150$ \\
4 & $79.117\pm0.683$ & $29.414\pm0.227$ & $29.414\pm0.227$ & $29.414\pm0.227$ & $29.414\pm0.227$ \\
5 & $113.162\pm0.997$ & $35.467\pm0.295$ & $35.467\pm0.295$ & $35.467\pm0.295$ & $35.467\pm0.295$ \\
6 & $161.263\pm1.486$ & $43.407\pm0.375$ & $43.407\pm0.375$ & $43.407\pm0.375$ & $43.407\pm0.375$ \\
7 & $210.521\pm2.230$ & $49.380\pm0.459$ & $49.380\pm0.459$ & $49.380\pm0.459$ & $49.380\pm0.459$ \\
\bottomrule
\end{tabular}

%% file: tables/results_assortment_timing_l2.tex
\begin{tabular}{@{}llrrrrrr@{}}
\toprule
 & & \multicolumn{4}{c}{Query} & \multicolumn{2}{c}{Solver runtime (s)} \\
\cmidrule(lr){3-6}\cmidrule(lr){7-8}
Family & Budget & Wall (s) & Sandbox (s) & Cells & Output tokens & Mean & Max \\
\midrule
MMNL & 3{,}600 s & 535 & 97.5 & 11 & 25,572 & 0.49 & 2.0 \\
Nested logit & 3{,}600 s & 464 & 121.3 & 9 & 19,889 & 0.75 & 2.9 \\
Constrained MMNL & 3{,}600 s & 708 & 98.6 & 11 & 36,119 & 0.98 & 25.4 \\
\midrule
MMNL & 0 s & 256 & 0.0 & 0 & 13,076 & 0.08 & 0.8 \\
Nested logit & 0 s & 256 & 0.0 & 0 & 12,631 & 0.40 & 1.6 \\
Constrained MMNL & 0 s & 259 & 0.0 & 0 & 13,900 & 0.09 & 2.0 \\
\bottomrule
\end{tabular}

%% file: tables/results_assortment_timing_l1.tex
\begin{tabular}{@{}lrrrrrrrrr@{}}
\toprule
 & \multicolumn{3}{c}{Wall (s)} & \multicolumn{3}{c}{Sandbox (s)} & Cells & \multicolumn{2}{c}{Output tokens} \\
\cmidrule(lr){2-4}\cmidrule(lr){5-7}\cmidrule(lr){9-10}
Family (\# queries) & Mean & SD & Max & Mean & SD & Max & Mean & Mean & SD \\
\midrule
MMNL (72) & 65 & 134 & 1053 & 26.7 & 115.7 & 891.7 & 3.8 & 1,627 & 1,145 \\
Nested logit (48) & 154 & 149 & 872 & 51.6 & 114.8 & 660.2 & 5.4 & 5,140 & 2,583 \\
Constrained MMNL (36) & 25 & 9 & 48 & 1.4 & 2.5 & 12.8 & 3.4 & 1,227 & 467 \\
\bottomrule
\end{tabular}

%% file: tables/results_l2_holdout_assortment.tex
\setlength{\tabcolsep}{4pt}
\begin{tabular}{@{}lcccc@{}}
\toprule
Class & \makecell{\model{gpt-5.6-sol}} & \makecell{\model{claude-fable-5}} & \makecell{\model{gpt-5.4}} & \makecell{\model{gpt-5.1}} \\
\midrule
\makecell[l]{MMNL exact\\ small (48)} &
\makecell{$1.0000$\\ $[1.0000,1.0000]$} &
\makecell{$1.0000$\\ $[1.0000,1.0000]$} &
\makecell{$1.0000$\\ $[1.0000,1.0000]$} &
\makecell{$1.0000$\\ $[1.0000,1.0000]$} \\
\makecell[l]{MMNL large\\ reference (16)} &
\makecell{$1.0000$\\ $[1.0000,1.0000]$} &
\makecell{$1.0000$\\ $[1.0000,1.0000]$} &
\makecell{$1.0000$\\ $[1.0000,1.0000]$} &
\makecell{$1.0000$\\ $[1.0000,1.0000]$} \\
\makecell[l]{Nested logit\\ anti-correlated (12)} &
\makecell{$1.0000$\\ $[1.0000,1.0000]$} &
\makecell{$1.0000$\\ $[1.0000,1.0000]$} &
\makecell{$1.0000$\\ $[1.0000,1.0000]$} &
\makecell{$1.0000$\\ $[1.0000,1.0000]$} \\
\makecell[l]{Nested logit\\ cross-nest (12)} &
\makecell{$1.0000$\\ $[1.0000,1.0000]$} &
\makecell{$1.0000$\\ $[1.0000,1.0000]$} &
\makecell{$1.0000$\\ $[1.0000,1.0000]$} &
\makecell{$1.0000$\\ $[1.0000,1.0000]$} \\
\bottomrule
\end{tabular}

%% file: prompts_appendix_examples.tex
\subsection{Inventory Prompt Examples}

\promptheading{Inventory, level 1 example: deterministic lost sales, lead time 8.}
\begin{lstlisting}[style=promptstyle]
You are asked to write an ordering policy for a SPECIFIC problem.

PROBLEM. A retailer manages a single product over an infinite horizon. Each period, in this order:
  (1) The order placed lead_time periods ago arrives and is added to on-hand inventory.
  (2) You place an order of size q, where 0 <= q <= max_order = 15. It ARRIVES AFTER A LEAD TIME of lead_time = 8 periods.
  (3) Demand for the period is realized. Demand is i.i.d. Poisson with KNOWN mean 5.
  (4) Sales = min(on_hand, demand). ANY UNMET DEMAND IS LOST (it is NOT backordered).
  (5) Costs for the period: holding_cost = 1 per unit of inventory left over after demand, plus penalty_cost = 4 per unit of demand that was lost. The ordering cost is 0.

Goal: choose an ordering policy that MINIMIZES the long-run average cost per period.
All parameters above are known exactly; nothing needs to be estimated.

You have a run_python(code) tool: it executes Python (numpy as np, and scipy, are available) and returns stdout. Preloaded variables: lead_time, demand_mean, penalty_cost, holding_cost, ordering_cost, max_order. State does NOT persist between calls; re-define what you need each call. print() whatever you need to see.

BUDGET. You have 3600 SECONDS OF TOTAL COMPUTATION across all run_python calls, to divide up however you like -- many short calls, a few long ones, or one long one. Each call may use up to whatever remains. After every call you will be told how much compute is left. You may make at most 50 calls. You may test and revise your approach as many times as you like.

When you are done, return ONLY the Python source of your final answer in a single ```python code block (plus any helper functions it needs), defining EXACTLY this signature:

    def order(on_hand, pipeline):

- on_hand: int, units on hand after this period's arrival; pipeline: tuple of length lead_time-1, pipeline[j] = quantity arriving in j+1 periods.
- It must return an integer order quantity in [0, max_order].
- The order(...) function is called once per simulated period and should be lightweight; it should not run simulations or expensive optimization inside each call.
- It need not be deterministic.
- Use only the Python standard library, numpy (as np), and scipy, plus the preloaded problem parameters. You may compute at module level to precompute tables or constants when your code loads; module-level precompute should run within 30 seconds.
- Everything outside that final code block is ignored. Do not wrap the code in JSON.
\end{lstlisting}

\promptheading{Inventory, level 2 example: deterministic lost-sales class.}
\begin{lstlisting}[style=promptstyle]
You are asked to write a general-purpose solver for a CLASS of problems.

PROBLEM CLASS. A retailer manages a single product over an infinite horizon. Each period, in this order:
  (1) The order placed lead_time periods ago arrives and is added to on-hand inventory.
  (2) An order of size q is placed, where 0 <= q <= max_order. It ARRIVES AFTER A LEAD TIME of lead_time periods.
  (3) Demand for the period is realized. Demand is i.i.d. Poisson with KNOWN mean demand_mean.
  (4) Sales = min(on_hand, demand). ANY UNMET DEMAND IS LOST (it is NOT backordered).
  (5) Costs for the period: holding_cost per unit of inventory left over after demand, plus penalty_cost per unit of demand that was lost, plus ordering_cost per unit ordered.

Goal: for each instance, produce an ordering policy that MINIMIZES that instance's long-run average cost per period. All parameters are known exactly; nothing needs to be estimated.

An INSTANCE of the class is a specific setting of the parameters (lead_time, demand_mean, holding_cost, penalty_cost, ordering_cost, max_order). Your solver will be run on MANY instances: lead_time in [2, 12], demand_mean up to 50, penalty_cost up to 100.

You have a run_python(code) tool: it executes Python (numpy as np, and scipy, are available) and returns stdout. No variables are preloaded and no instances from the evaluation set are provided. State does NOT persist between calls; re-define what you need each call. print() whatever you need to see.

BUDGET. You have 3600 SECONDS OF TOTAL COMPUTATION across all run_python calls, to divide up however you like -- many short calls, a few long ones, or one long one. Each call may use up to whatever remains. After every call you will be told how much compute is left. You may make at most 50 calls. You may test and revise your approach as many times as you like.

When you are done, return ONLY the Python source of your final answer in a single ```python code block (plus any helper functions it needs), defining EXACTLY this signature:

    def design(lead_time, demand_mean, holding_cost, penalty_cost, ordering_cost, max_order):

- design(...) receives the six instance parameters and returns a FUNCTION order(on_hand, pipeline): on_hand is an int (units on hand after this period's arrival); pipeline is a tuple of length lead_time-1, pipeline[j] = quantity arriving in j+1 periods; it must return an integer order quantity in [0, max_order].
- design() may precompute whatever it needs (tables, constants). When your solver is used, design(...) should run within 30 seconds per instance.
- The returned order(...) function is called once per simulated period and should be lightweight; it should not run simulations or expensive optimization inside each call.
- It need not be deterministic.
- Use only the Python standard library, numpy (as np), and scipy.
- It must never raise for any valid input satisfying the problem description and shape/range constraints above.
- Everything outside that final code block is ignored. Do not wrap the code in JSON.
\end{lstlisting}

\subsection{Queueing Prompt Examples}

\promptheading{Queueing, level 1 example: reentrant-line network with $L=2$ stations.}
\begin{lstlisting}[style=promptstyle]
You are asked to write a scheduling policy for a SPECIFIC problem.

PROBLEM. Scheduling in a multiclass queueing network with S servers and Q queues (job classes), in continuous time.
  * Jobs arrive to queue q from outside according to a Poisson process with rate lam[q] (lam[q] may be 0).
  * Server s can work on queue q if and only if network[s][q] == 1. Every job carries a service requirement, drawn when it joins a queue independently of everything else from an exponential distribution with mean 1; a server working on a job in queue q depletes its requirement at rate mu[s][q], and the job completes when the requirement reaches 0. So a job in queue q served without interruption by server s takes an Exp(mu[s][q]) time.
  * When a job completes at queue q it moves to queue next_queue[q], where it draws a new service requirement, or leaves the system if next_queue[q] == -1.
  * Each job in queue q costs h[q] per unit time while it is there.
  * A server works on at most one job at a time and a job is worked on by at most one server at a time. Service is preemptive-resume: at every event (an arrival or a service completion) the controller observes the vector of queue lengths and the current time and decides which queue each server works on until the next event; a server may be moved to a different queue at any event, and an interrupted job keeps its remaining requirement. Several servers sent to the same queue work on distinct jobs in order of arrival (the faster server takes the older job). A server may also be left idle.

Goal: produce a scheduling policy that MINIMIZES the long-run (steady-state) average holding cost per unit time, sum_q h[q] * (number of jobs in queue q). All parameters are known exactly; nothing needs to be estimated.

THE INSTANCE. S = 2 servers, Q = 6 queues.
  network = [[1, 1, 1, 0, 0, 0], [0, 0, 0, 1, 1, 1]]
  mu = [[0.125, 0.5, 0.25, 0, 0, 0], [0, 0, 0, 0.166667, 0.142857, 1]]
  lam = [0.0642857, 0, 0.0642857, 0, 0, 0]
  next_queue = [3, 4, 5, 1, -1, -1]
  h = [1, 1, 1, 1, 1, 1]
These are preloaded in the sandbox as Python lists (network, mu, lam, next_queue, h) and the ints S and Q.

You have a run_python(code) tool: it executes Python (numpy as np, and scipy, are available) and returns stdout. Preloaded variables: network, mu, lam, next_queue, h, S, Q. State does NOT persist between calls; re-define what you need each call. print() whatever you need to see.

BUDGET. You have 3600 SECONDS OF TOTAL COMPUTATION across all run_python calls, to divide up however you like -- many short calls, a few long ones, or one long one. Each call may use up to whatever remains. After every call you will be told how much compute is left. You may make at most 50 calls. You may test and revise your approach as many times as you like.

When you are done, return ONLY the Python source of your final answer in a single ```python code block (plus any helper functions it needs), defining EXACTLY this signature:

    def policy(queues, t):

- queues is a (Q,) integer ndarray of the current queue lengths and t is the current time; policy(...) must return a sequence of S ints, entry s being the index of the queue server s works on until the next event, or -1 to idle. The chosen queue must satisfy network[s][q] == 1, and no more servers may be sent to a queue than it has jobs.
- The policy(...) function is called once per event (tens of thousands of times per simulated trajectory, many trajectories) and must be lightweight, typically well under a millisecond per call; it should not run simulations or expensive optimization inside each call. You may compute at module level to precompute tables or constants when your code loads; module-level precompute should run within 30 seconds.
- It need not be deterministic.
- Use only the Python standard library, numpy (as np), and scipy, plus the preloaded instance variables.
- It must never raise for any valid input.
- Everything outside that final code block is ignored. Do not wrap the code in JSON.
\end{lstlisting}

\promptheading{Queueing, level 2 example: criss-cross class.}
\begin{lstlisting}[style=promptstyle]
You are asked to write a general-purpose solver for a CLASS of problems.

PROBLEM CLASS. Scheduling in a queueing network with a FIXED topology, in continuous time: S = 2 servers and Q = 3 queues, with
    network = [[1, 0, 1], [0, 1, 0]]   next_queue = [1, -1, -1]   h = [1, 1, 1]
That is: jobs arrive from outside to queue 0 and to queue 2 (Poisson, rates lam[0] and lam[2]; lam[1] = 0). Server 0 can work on queues 0 and 2. When a queue-0 job completes it moves to queue 1, which only server 1 can work on; queue-1 and queue-2 jobs leave the system on completion. Every job costs 1 per unit time while it is in the system.

  * Server s can work on queue q if and only if network[s][q] == 1. Every job carries a service requirement, drawn when it joins a queue independently of everything else from an exponential distribution with mean 1; a server working on a job in queue q depletes its requirement at rate mu[s][q], and the job completes when the requirement reaches 0. So a job in queue q served without interruption by server s takes an Exp(mu[s][q]) time.
  * When a job completes at queue q it moves to queue next_queue[q], where it draws a new service requirement, or leaves the system if next_queue[q] == -1.
  * Each job in queue q costs h[q] per unit time while it is there.
  * A server works on at most one job at a time and a job is worked on by at most one server at a time. Service is preemptive-resume: at every event (an arrival or a service completion) the controller observes the vector of queue lengths and the current time and decides which queue each server works on until the next event; a server may be moved to a different queue at any event, and an interrupted job keeps its remaining requirement. Several servers sent to the same queue work on distinct jobs in order of arrival (the faster server takes the older job). A server may also be left idle.

Goal: for each instance, produce a scheduling policy that MINIMIZES the long-run (steady-state) average holding cost per unit time, sum_q h[q] * (number of jobs in queue q). All parameters are known exactly; nothing needs to be estimated.

An INSTANCE of the class is a specific setting of the rates (lam[0], lam[2], mu[0][0], mu[1][1], mu[0][2]); the topology above never changes. Your solver will be run on MANY instances, from lightly loaded to heavily loaded (server loads up to 0.9, always stable). The arguments are passed in the same format as stated below.

You have a run_python(code) tool: it executes Python (numpy as np, and scipy, are available) and returns stdout. No variables are preloaded and no instances from the evaluation set are provided. State does NOT persist between calls; re-define what you need each call. print() whatever you need to see.

BUDGET. You have 3600 SECONDS OF TOTAL COMPUTATION across all run_python calls, to divide up however you like -- many short calls, a few long ones, or one long one. Each call may use up to whatever remains. After every call you will be told how much compute is left. You may make at most 50 calls. You may test and revise your approach as many times as you like.

When you are done, return ONLY the Python source of your final answer in a single ```python code block (plus any helper functions it needs), defining EXACTLY this signature:

    def design(network, mu, lam, next_queue, h):

- network: (S,Q) 0/1 ndarray; mu: (S,Q) ndarray with mu[s][q] > 0 iff network[s][q] == 1; lam: (Q,) ndarray; next_queue: (Q,) int ndarray, -1 meaning the job leaves; h: (Q,) ndarray.
- design(...) returns a FUNCTION policy(queues, t): queues is a (Q,) integer ndarray of the current queue lengths and t is the current time; it must return a sequence of S ints, entry s being the index of the queue server s works on until the next event, or -1 to idle. The chosen queue must satisfy network[s][q] == 1, and no more servers may be sent to a queue than it has jobs.
- design() may precompute whatever it needs (tables, constants). When your solver is used, design(...) should run within 30 seconds per instance.
- The returned policy(...) function is called once per event (tens of thousands of times per simulated trajectory, many trajectories per instance) and must be lightweight, typically well under a millisecond per call; it should not run simulations or expensive optimization inside each call.
- It need not be deterministic.
- Use only the Python standard library, numpy (as np), and scipy.
- It must never raise for any valid input satisfying the problem description and shape/range constraints above.
- Everything outside that final code block is ignored. Do not wrap the code in JSON.
\end{lstlisting}

\promptheading{Queueing, Broad L2 example: multiclass-network class.}
\begin{lstlisting}[style=promptstyle]
You are asked to write a general-purpose solver for a CLASS of problems.

PROBLEM CLASS. Scheduling in a multiclass queueing network with S servers and Q queues (job classes), in continuous time.
  * Jobs arrive to queue q from outside according to a Poisson process with rate lam[q] (lam[q] may be 0).
  * Server s can work on queue q if and only if network[s][q] == 1. Every job carries a service requirement, drawn when it joins a queue independently of everything else from an exponential distribution with mean 1; a server working on a job in queue q depletes its requirement at rate mu[s][q], and the job completes when the requirement reaches 0. So a job in queue q served without interruption by server s takes an Exp(mu[s][q]) time.
  * When a job completes at queue q it moves to queue next_queue[q], where it draws a new service requirement, or leaves the system if next_queue[q] == -1.
  * Each job in queue q costs h[q] per unit time while it is there.
  * A server works on at most one job at a time and a job is worked on by at most one server at a time. Service is preemptive-resume: at every event (an arrival or a service completion) the controller observes the vector of queue lengths and the current time and decides which queue each server works on until the next event; a server may be moved to a different queue at any event, and an interrupted job keeps its remaining requirement. Several servers sent to the same queue work on distinct jobs in order of arrival (the faster server takes the older job). A server may also be left idle.

Goal: for each instance, produce a scheduling policy that MINIMIZES the long-run (steady-state) average holding cost per unit time, sum_q h[q] * (number of jobs in queue q). All parameters are known exactly; nothing needs to be estimated.

An INSTANCE of the class is a specific setting of (network, mu, lam, next_queue, h). Your solver will be run on MANY instances: up to 7 servers and 21 queues, stable but possibly heavily loaded.

You have a run_python(code) tool: it executes Python (numpy as np, and scipy, are available) and returns stdout. No variables are preloaded and no instances from the evaluation set are provided. State does NOT persist between calls; re-define what you need each call. print() whatever you need to see.

BUDGET. You have 3600 SECONDS OF TOTAL COMPUTATION across all run_python calls, to divide up however you like -- many short calls, a few long ones, or one long one. Each call may use up to whatever remains. After every call you will be told how much compute is left. You may make at most 50 calls. You may test and revise your approach as many times as you like.

When you are done, return ONLY the Python source of your final answer in a single ```python code block (plus any helper functions it needs), defining EXACTLY this signature:

    def design(network, mu, lam, next_queue, h):

- network: (S,Q) 0/1 ndarray; mu: (S,Q) ndarray with mu[s][q] > 0 iff network[s][q] == 1; lam: (Q,) ndarray; next_queue: (Q,) int ndarray, -1 meaning the job leaves; h: (Q,) ndarray.
- design(...) returns a FUNCTION policy(queues, t): queues is a (Q,) integer ndarray of the current queue lengths and t is the current time; it must return a sequence of S ints, entry s being the index of the queue server s works on until the next event, or -1 to idle. The chosen queue must satisfy network[s][q] == 1, and no more servers may be sent to a queue than it has jobs.
- design() may precompute whatever it needs (tables, constants). When your solver is used, design(...) should run within 30 seconds per instance.
- The returned policy(...) function is called once per event (tens of thousands of times per simulated trajectory, many trajectories per instance) and must be lightweight, typically well under a millisecond per call; it should not run simulations or expensive optimization inside each call.
- It need not be deterministic.
- Use only the Python standard library, numpy (as np), and scipy.
- It must never raise for any valid input satisfying the problem description and shape/range constraints above.
- Everything outside that final code block is ignored. Do not wrap the code in JSON.
\end{lstlisting}

\subsection{Assortment Prompt Examples}

\promptheading{Assortment, level 1 example: MMNL.}
\begin{lstlisting}[style=promptstyle]
You are asked to solve ONE SPECIFIC instance of an optimization problem. The full instance data is preloaded in your Python tool.

PROBLEM CLASS. Assortment optimization under the MIXED MULTINOMIAL LOGIT (MMNL) choice model
(also called the latent-class or mixture-of-logits model).

A retailer offers a subset S of n products. Customers belong to m segments; segment j occurs with
probability omega[j]. A customer in segment j chooses product i in S with probability
    u[j][i] / (v0[j] + sum_{k in S} u[j][k])
and buys nothing with probability v0[j] / (v0[j] + sum_{k in S} u[j][k]).
Product i earns price[i]. Expected revenue is

    R(S) = sum_j omega[j] * ( sum_{i in S} price[i]*u[j][i] ) / ( v0[j] + sum_{i in S} u[j][i] )

Goal: choose S maximizing R(S), subject to |S| <= cap when cap is not None.

THIS INSTANCE: n = 200, m = 25. The instance variables are preloaded in the sandbox exactly as defined above.

You have a run_python(code) tool: it executes Python (numpy as np, and scipy, are available) and returns stdout. Preloaded variables: u, price, v0, omega, cap, n, m. State does NOT persist between calls; re-define what you need each call. print() whatever you need to see.

BUDGET. You have 900 SECONDS OF TOTAL COMPUTATION across all run_python calls, to divide up however you like -- many short calls, a few long ones, or one long one. Each call may use up to whatever remains. After every call you will be told how much compute is left. You may make at most 50 calls. You may test and revise your approach as many times as you like.

When you are done, give your FINAL ANSWER as the last line of your reply, in the form:

    FINAL: [i1, i2, ...] -- the offered product indices as a Python list

- The FINAL line must be valid Python literal syntax, nothing else on the line.
- Return a list of distinct product indices (|S| <= cap when cap is not None). Indices are 0-based.
- Everything before the FINAL line is ignored. Exactly one FINAL line.
\end{lstlisting}

\promptheading{Assortment, level 2 example: MMNL class.}
\begin{lstlisting}[style=promptstyle]
You are asked to write a general-purpose solver for a CLASS of problems.

PROBLEM CLASS. Assortment optimization under the MIXED MULTINOMIAL LOGIT (MMNL) choice model
(also called the latent-class or mixture-of-logits model).

A retailer offers a subset S of n products. Customers belong to m segments; segment j occurs with
probability omega[j]. A customer in segment j chooses product i in S with probability
    u[j][i] / (v0[j] + sum_{k in S} u[j][k])
and buys nothing with probability v0[j] / (v0[j] + sum_{k in S} u[j][k]).
Product i earns price[i]. Expected revenue is

    R(S) = sum_j omega[j] * ( sum_{i in S} price[i]*u[j][i] ) / ( v0[j] + sum_{i in S} u[j][i] )

Goal: choose S maximizing R(S), subject to |S| <= cap when cap is not None.

An INSTANCE of the class is a specific setting of (u, price, v0, omega, cap). Your solver will be run on MANY instances with n up to 200 and m up to 25.

You have a run_python(code) tool: it executes Python (numpy as np, and scipy, are available) and returns stdout. No variables are preloaded and no instances from the evaluation set are provided. State does NOT persist between calls; re-define what you need each call. print() whatever you need to see.

BUDGET. You have 3600 SECONDS OF TOTAL COMPUTATION across all run_python calls, to divide up however you like -- many short calls, a few long ones, or one long one. Each call may use up to whatever remains. After every call you will be told how much compute is left. You may make at most 50 calls. You may test and revise your approach as many times as you like.

When you are done, return ONLY the Python source of your final answer in a single ```python code block (plus any helper functions it needs), defining EXACTLY this signature:

    def solve(u, price, v0, omega, cap):

- u: (m,n) ndarray; price: (n,) ndarray; v0: (m,) ndarray; omega: (m,) ndarray; cap: int or None.
- Return a list of distinct product indices (|S| <= cap when cap is not None). Indices are 0-based.
- When your solver is used, solve(...) should run within 30 seconds per instance.
- It must never raise for any valid input satisfying the problem description and shape/range constraints above. Use only the Python standard library, numpy (as np), and scipy.
- Everything outside that final code block is ignored. Do not wrap the code in JSON.
\end{lstlisting}